\documentclass{article}
\usepackage{blindtext}
\usepackage[english]{babel}
\usepackage{amsmath}
\usepackage{url}
\usepackage{graphicx}
\usepackage[final]{pdfpages}
\graphicspath{{./imgs/}}
\usepackage[bottom,marginal,ragged]{footmisc}

\usepackage{float}
\usepackage[parfill]{parskip}
\usepackage{graphicx}
\usepackage{tikz}
\usepackage{csquotes}
\usepackage{tcolorbox}
\usepackage[margin=0.7in]{geometry}
\usepackage{multicol}
\usepackage{mathtools}
\usetikzlibrary{patterns}
\DeclareRobustCommand{\legendsquare}[1]{%
  \tikz[baseline=(a.south)]{\node[#1, inner sep=.8ex, outer sep=0] (a) {};}%
}
\usepackage[none]{hyphenat}
\definecolor{ConvOrange}{RGB}{255,234,193}
\definecolor{ReLUOrange}{RGB}{255,194,128}
\definecolor{MaxRed}{RGB}{225,97,63}
\definecolor{UpBlue}{RGB}{178,203,229}
\definecolor{SoftPurple}{RGB}{135,38,109}

\begin{document}
\pagenumbering{roman}
\begin{titlepage}
   \centering
   \rule{\textwidth}{1pt}\\[\baselineskip]
   \huge{Heatmap-based Object Detection and Tracking \linebreak with a Fully Convolutional Neural Network}\\
   \rule{\textwidth}{1pt}\vspace*{-\baselineskip}\vspace{1.5cm}
   \begin{minipage}{0.47\textwidth}
      \begin{flushleft} 
         \large{\textbf{Fabian Amherd}}\\
         \large{\textbf{Elias Rodriguez}}
      \end{flushleft}
      \end{minipage}
      \begin{minipage}{0.47\textwidth}
      \begin{flushright} \large
          \large{fabian.amherd@gmail.com}
          \large{elias.rodriguez@bluewin.ch}
      \end{flushright}
   \end{minipage}\\
   \vspace{1.5cm}
   \begin{minipage}{0.47\textwidth}
      \begin{flushleft} 
         \large{\textbf{Advisors:}\\Michael Bucher\\Christopher Latkoczy}
      \end{flushleft}
      \end{minipage}
      \begin{minipage}{0.47\textwidth}
      \begin{flushright}         
         \large{\textbf{Kantonsschule Stadelhofen}\\Schanzengasse 17\\8001 Zürich, Switzerland}
      \end{flushright}
   \end{minipage}\\
   \vspace{1.5cm}
   \begin{minipage}{0.47\textwidth}
      \begin{flushleft} 
         \large{\textbf{External Advisors:}\\Prof. Dr. Tobias Delbrück\\Yuhuang Hu\\Shasha Guo}
      \end{flushleft}
      \end{minipage}
      \begin{minipage}{0.47\textwidth}
      \begin{flushright}         
         \large{\textbf{Institute of Neuroinformatics}\\Winterthurerstrasse 190\\8057 Zürich, Switzerland}
      \end{flushright}
   \end{minipage}
   \vfill
   \large{\the\year}\\
\end{titlepage}

\pagebreak
\vspace*{\fill}
\begin{center}
   \textit{Per Aspera Ad Astra}
\end{center}
\vspace*{\fill}
\pagebreak
\tableofcontents
\pagebreak

\begin{multicols*}{2}
\pagenumbering{arabic}
\section*{Abstract}
\textbf{The main topic of this paper is a brief overview of the field of Artificial Intelligence. The core of this paper is a practical implementation of an algorithm for object detection and tracking. The ability to detect and track fast-moving objects is crucial for various applications of Artificial Intelligence like autonomous driving, ball tracking in sports, robotics or object counting. As part of this paper the Fully Convolutional Neural Network "CueNet" was developed. It detects and tracks the cueball on a labyrinth game robustly and reliably. While CueNet V1 has a single input image, the approach with CueNet V2 was to take three consecutive 240 x 180-pixel images as an input and transform them into a probability heatmap for the cueball's location. The network was tested with a separate video that contained all sorts of distractions to test its robustness. When confronted with our testing data, CueNet V1 predicted the correct cueball location in 99.6\% of all frames, while CueNet V2 had 99.8\% accuracy. To get access to the source code of the project, scan the QR-code at the beginning of chapter \ref{practicalpart}.}

\section{Introduction}
\subsection{What is Artificial Intelligence?}
\subsubsection{\mbox{Definition of Artificial Intelligence (AI)}}

\vspace{5mm}
\begin{displayquote}
   \textit{Artificial Intelligence (AI) is the simulation of intelligence in a non-biological structure.}
\end{displayquote}
\vspace{5mm}

As in many fields, there does not exist one single agreed-upon definition of Artificial Intelligence. A definition often used is "Artificial intelligence (AI) would be the possession of intelligence, or the exercise of thought, by machines such as computers." \cite{iep}. So the only difference to Human Intelligence would be, that the intelligence is possessed by a machine on a semiconductor-basis, not by a human in a biological structure.\\
The question "Can machines think?" is very interesting to consider here.
One can also have the opinion that a machine cannot possess real Intelligence, that it can only simulate intelligent behaviour. Therefore, another definition that is around, reads as follows: "AI [can be looked at] as an engineering discipline in which researchers focus on developing useful programs and tools that perform in domains that normally require intelligence." \cite{sd}. So we can also define Artificial Intelligence as algorithms able to solve tasks, that normally require intelligence to be solved, although these algorithms do not possess "Intelligence" in a traditional sense.
\subsubsection{Different types of AI}
Some people may find it useful to put different algorithms or "AI systems" into categories, as a criterion for which can be used the comparison to a human. The following categorization is based on the capability of an AI. There is another popular categorization based on functionality, which won't be considered here though.\\

The least sophisticated type of AI is called \textbf{Artificial Narrow Intelligence (ANI)} or "weak AI". This is where we would place all of the AI systems we have encountered up to this point. As Forbes puts it: "Artificial narrow intelligence refers to AI systems that can only perform a specific task autonomously using human-like capabilities." \cite{forbestypes}. Important to consider here is that these systems can be operating at a superhuman level of performance, but only in one specific area they were programmed for. \\
One step up the ladder we have \textbf{Artificial General Intelligence (AGI)} or "strong AI". These AI systems have human-level intelligence across all domains, they are as multi-functional as humans. They can understand, discuss and generalize like human beings. "AGI can think, understand, and act in a way that is indistinguishable from that of a human in any given situation." \cite{ttypes} \\
An \textbf{Artificial Superintelligence (ASI)} would surpass the capabilities of humans in every aspect. It would be able to come up with ideas that are impossible to grasp for humans. Whether it is in science, sports, understanding emotions, creative thinking, ASI will outperform humans. The potential of such a machine is tremendous and comes with unknown consequences. \\
It has been proposed that the emergence of ASI is almost inevitable when AGI has been achieved. To justify this, one only has to look at the advantages of a digital implementation over a biological one. In the digital implementation of a brain, the simulation can be copied and accelerated multiple times. Thus, when we create AGI with a whole brain emulation, we can, with sufficient computational resources, create a collection of such AIs that work together at superhuman speeds. \cite{tts15}

\subsubsection{A closer look at the field of AI}
\subsubsection*{Machine Learning (ML)} \label{ML}

\vspace{5mm}
\begin{displayquote} 
   \textit{Machine Learning (ML) is the process that an Artificial Intelligence goes through when learning.}
\end{displayquote}
\vspace{5mm}

\begin{figure}[H]
   \begin{center}
      \includegraphics[width=.45\textwidth]{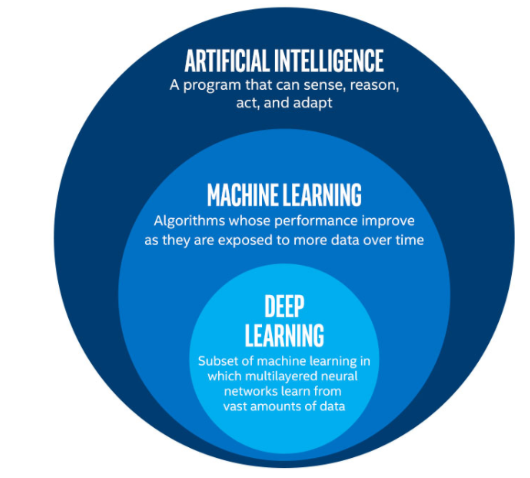}
   \end{center}
   \caption[Source: https://towardsdatascience.com/what-is-deep-learning-and-how-does-it-work-2ce44bb692ac]{AI, Machine Learning and Deep Learning in a hierarchical order.}
   \label{fig:AIMLDL}
\end{figure}

Another term for Artificial Intelligence, although less used, is "Machine Intelligence". "Machine Learning" is what leads to it.\\
Machine Learning is the process that an algorithm goes through when trying to search for patterns in a massive amount of data. The term refers to a subfield of AI (see figure \ref{fig:AIMLDL}), but still contains numerous other subfields (one of which is Deep Learning). The big advantage of Machine Learning is that the algorithm tries to find the patterns automatically, it learns through the experience it gets from the data that gets fed into it. Mathematically, the program tries to find the best possible function to map a certain input to a certain output. This is also why neural networks are sometimes called "universal function approximators".\\
There are different approaches to how a programmer can set up the environment for an algorithm to learn. One approach is called \textbf{Supervised Learning}, its principle is to present the algorithm with the input, let it output something, and then give feedback on how close the actual output was to the desired output, also called the "Loss", by calculating the difference between the actual output and the label. There are different ways to calculate this difference, which will be discussed in more detail in \ref{lossfn} of this paper. \\
A second approach often used is called \textbf{Unsupervised Learning}. The goal here is to let the algorithm automatically find the underlying structure of the data fed into it. This may help to discover connections that are not recognizable by humans alone because we can't take as much data into respect as computers can. \\
A third popular approach is \textbf{Reinforcement Learning}. Here the algorithm becomes an "agent" in a simulated or real environment in which it can take actions that lead to a different state of the environment. The algorithm constantly tries to change its "policy" of taking actions to maximize the reward it gets. \cite{mitml} \\

\subsubsection*{Deep Learning (DL)}
\vspace{5mm}
\begin{displayquote} 
   \textit{If we perform Machine Learning (ML) with a Deep Neural Network (a form of an AI), it is called Deep Learning (DL).}
\end{displayquote}
\vspace{5mm}
One of the biggest subfields of Machine Learning is named "Deep Learning". This field addresses the development of \textbf{Artificial Neural Networks (ANNs)}, whose structure is strongly inspired by the human brain. ANNs will be discussed in more detail in the first part of \ref{ANNs} . More complex tasks often require bigger networks with more subsequent layers. If an ANN consists of more than three layers, we call it a \textbf{Deep Neural Network}.\cite{mlmastery} \\
There are dozens of different architectures of ANNs that excel at specific types of problems. The architectures we will have a closer look at in this paper are the most simple form, the Dense Neural Network (DNN), as well as the Convolutional Neural Network (CNN), which is being used in the practical part of this work.\\
Deep Learning is already being applied in all kinds of different industries all over the world, most widely known are recommendation systems as on Youtube or Amazon. Although already used so much, it is said to still have enormous potential to transform our lives completely. The realised and potential applications of AI (and with it Deep Learning) will be discussed more closely in \ref{app} of this paper.

\subsection{History}
The history of AI dates way back several thousand years where myths and stories of artificial beings with intelligence were being told. Amongst the old greeks there existed the myth of Talos, a guardian of the island of Crete, which was artificially forged by Hephaestus with the aid of a cyclops. His task was to throw boulders at ships of invaders and walk three times around the island's perimeter daily \cite{aihistory}. From 380 BC to 1900 there have been numerous mathematicians, theologians, philosophers, professors and authors who mused about calculating machines, mechanical techniques and other systems that eventually led to the concept of artificial human-like thought in non-human beings.

\subsubsection{Earlier Epochs}
The first real implementation of an AI was proposed by Warren McCulloch and Walter Pitts in 1943 with a mathematical model, combined with a computer model of the biological neuron. \cite{historybritannica}
At the beginning of 1950, Alan Turing and John von Neumann became the founding fathers of the technology behind AI, as they made the transition from analogue and hard-wired digital computers to computers with a stored program in a separate memory (the so-called "von Neumann" architecture). Turing and Neumann thus formalized the architecture of the computers from today. Turing raised the question of the possible intelligence of a machine in his famous article "Computing Machinery and Intelligence" from 1950. In this article, he described a "game of imitation" which later was called the "Turing test", where the machine can pass it if a human moderator can not tell if he is talking to a human or a machine. \cite{aihistory} The field was in its golden years, people were extremely optimistic, claiming that within ten years a digital computer will be the world's first chess champion and within ten more years a digital computer will discover and prove an important new mathematical theorem. \cite{aioptimism}

\subsubsection*{The first AI winter 1974-1980}
AI remained fascinating but the popularity of the technology fell back in the early 1970s. AI would become subject to critiques and financial setbacks. The tremendous optimism had raised expectation impossibly high, which lead to the unfulfillment of the promises. \cite{aiwinter1} Combined with the limitations of the hardware at that time, on top of which came the cutting off from almost all funding for undirected research into AI, the technology was lead into its first winter. During the same period, Marvin Misky harshly criticised perceptrons and therefore shut down the field of neural nets almost completely for the next 10 years. \cite{criticwinter1} 

\subsubsection*{Boom 1980-1987}
In the 1980s, a form of AI called expert systems was adopted by corporations around the world. These expert systems are programs that can answer questions or solve problems about a specific domain of knowledge. An expert system tries to emulate the decision-making ability of a human expert, hence the name, by going through many "if–then" rules. The earliest examples were developed by Edward Feigenbaum and his students \cite{expertsystems1}\cite{expertsystems2}. Expert systems only function in a specific field of knowledge and all in all were useful, which AI had not been able to achieve up to this point.  \cite{expertsystems3}\cite{expertsystems4}

During the same years in Japan, the government started to fund AI with its fifth-generation computer project. The Japanese Ministry set aside \$850 million for the project. The goal was to create programs and machines that could do things like translate language and reason like humans. \cite{jpfunding}
As a reaction, other countries started to fund new projects in the field of AI too. In the USA a few companies came together to form the Microelectronics and Computer Technology Corporation for funding large projects in the field of AI. The Alvey project started in the UK which cost  £350 million.  \cite{usafunding}

\subsubsection*{The second AI winter 1987–1993}

In a similar pattern as the first winter, the second one occurred. Expectations were again set much too high and AI was exposed to financial setbacks. One could compare it to an economic bubble. Systems for vision and speech worked very poorly with expert systems because they contained too many edge cases, which couldn't be manually designed around. \cite{secondwinter}
\subsubsection*{AI 1993–2011}

Now some of the initial goals of AI could finally be achieved. May 11th 1997, Deep Blue, a chess-playing computer won a game against the world chess champion Garry Kasparov \cite{deepblue}. AI's success started to spread throughout the technology industry.
The overall success of AI was not due to some revolutionary discovery, but mostly due to the tremendous increase in the speed and capacity of computers in the 90s \cite{speed}. This dramatic increase is described by Moore's law, which states that the density of transistors on an Integrated Circuit (IC) of new computers doubles about every two years.
\subsubsection{AI today}
\subsubsection*{2011-present}
Since the beginning of the 2010s, the computer science branch of AI has experienced a boom (again). The reason is mainly the availability of three key resources in AI: Computing power, the internet and data. The latter is available in such quantities that it has even gotten a name: "\textbf{Big Data}".\\
These circumstances also led to significantly increased budgets for AI R\&D in the industry, the results of which we could observe in the last few years: In 2011, IBMs Watson won the TV-Show "Jeopardy", in 2016, AlphaGO beat the best human players, in 2020, Agent 57 beat the human benchmark in all 57 Atari games \cite{badia2020agent57} and the list of new achievements goes on. \cite{aihistory}\\
\subsubsection*{"Who is Who" in the field of AI?} \label{TPUs}
The point of this section is to give a quick overview of the most important players in the AI Industry regarding the big technology companies. \\

\textbf{Nvidia} is a company based in California, USA, that, among other things, develops Graphics Processing Units (GPUs). GPUs allow for massively parallel computation, and therefore, accelerated training of neural networks. As of Q4 in 2019, Nvidia has a GPU-shipment market share of $\sim$70\% \cite{nvidms}. \cite{nvidwiki} Nvidia is mostly fueled by the video game industry as heavy parallel computing power is required to run the latest video games.\\
\textbf{Google} is one of the major players in the AI industry. Google launched a separate department called "Google AI", that does R\&D. Waymo, a company owned by Google, currently has one of the most advanced autonomous driving systems. In 2016, Google introduced the Tensor Processing Unit (TPU) (See chapter \ref{tensors}). TPUs are integrated circuits specifically designed for machine learning with neural networks \cite{tpu}. Of course, there is also the "Google Assistant", which is a virtual assistant that can take your language or keyboard input, and search the internet for answers (amongst other things). Google has even launched a machine learning platform called "Tensorflow". \\
At \textbf{IBM}, after breakthroughs like DeepBlue in 1996/1997 and Watson in 2011, "Project Debater" has been in development since 2012 and has already debated against the debating world-champion Harish Natarajan in 2019, though it has lost that discussion. IBM also has an R\&D department called "IBM Research" that, amongst other things, does research at the cutting edge of AI. \cite{projectdebater} \\
Just like Google and Amazon, \textbf{Microsoft} has its own Cloud Computing platform. It is called "Microsoft Azure" and is partly devoted to run heavy AI workloads. Microsoft is building AI into their existing products such as Office 365 to make smart recommendations. One of Microsoft's initiatives is called "AI for Good", with which it tries to have a positive impact on the present and future development of AI. \\
At \textbf{Amazon}, AI is not located in a single department or office. Instead, AI is distributed and applied throughout the whole company. The most apparent application is the recommendation engine on the company's online shop. Similar to Google, Amazon developed a virtual assistant called "Alexa". One major advantage for Amazon is its access to a huge data source from tracking the purchase behaviour of their clients. \cite{amznai} \\
Like Google, \textbf{Facebook} has developed a Machine Learning library called "Pytorch", which is popular amongst researchers. Facebook also has a research department called "Facebook AI Research (FAIR)", that does research as well as publish useful tools like "Detectron 2", a platform that includes several implementations or different computer vision algorithm, including object detectors like the one of our study. \cite{fbai}\\
\vfill\null
\columnbreak
\section{Practical Part} \label{practicalpart}
This paper includes a practical project with a real-life application of an AI. This following chapter is devoted to the documentation of the experiment, as well as some theoretical explanations on the side to better illustrate the mathematics behind neural networks. The code is available under the following GitHub repository: \url{https://github.com/FabianAmherd/CNN_ER_FA} (see QR-code below).
\begin{figure}[H]
   \centering
   \includegraphics[width=.25\textwidth]{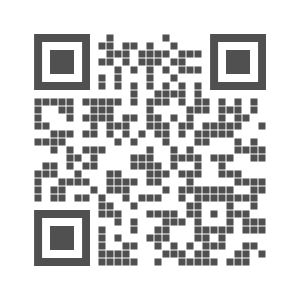}
\end{figure}
\subsection{Experiment} \label{exp}
\begin{figure}[H]
   \centering
   \includegraphics[width=.35\textwidth]{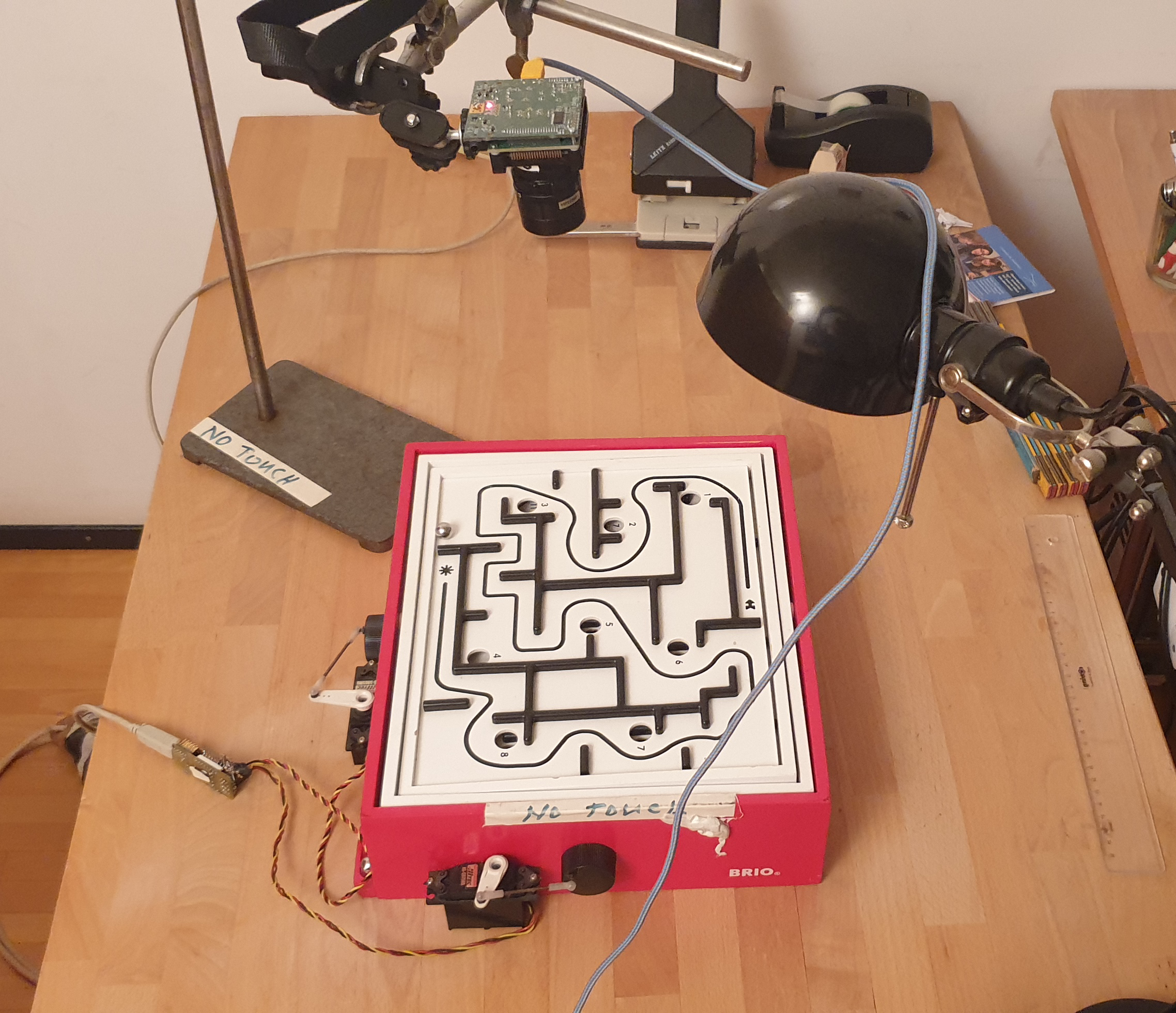}
   \caption[Source: Image created by authors of this paper.]{Setup: Brio labyrinth with RC servo motors, DAVIS camera and the labelling software.}
   \label{fig:setup}
\end{figure}

The challenge of the labyrinth game in figure \ref{fig:setup} is to roll the ball to a destination without it falling into one of the holes on the board. The ball is moved by tilting the labyrinth with two knobs (one on each axis). 
Our given task was to develop a neural network to detect the ball on this labyrinth game for tracking it. Specifically, we developed a fully convolutional neural network. With that, the first step for a fully automated labyrinth game was developed. The data was collected and labelled by ourselves with a camera and suitable labelling software.

\subsection{Materials and Methods}
The following section contains explanations and documentation to how the task described in \ref{exp} was solved. Also, it will cover the theoretical background of some popular Deep Learning methods or tools.
\subsubsection*{Equipment}
\begin{itemize}
   \item \textbf{Camera:} The DAVIS camera used is a dual-frame and event camera \cite{Brandli2014-fj}. It outputs conventional frames and a stream of dynamic vision sensor (DVS) brightness change events \cite{Lichtsteiner2008-km}. The events can be collected to variable duration frames to drive a conventional CNN with sparse frame input, allowing tracking of fast-moving objects.
   \item \textbf{Camera/Labelling Software:} To record and label our data, the project has used jAER software \cite{jaer} as well as the program "After Effects".
   \item \textbf{Labyrinth:} The labyrinth used was a "Brio Labyrinth Game" \cite{labi}. It is fitted with a camera stand as well as two RC servo motors to control the table tilt. These servo motors can be computer-controlled from a computer using the jAER software.
   \item \textbf{Machine Learning tools:} The project has used the open-source machine learning library \textbf{Pytorch} for developing, training and testing the CNN.
   \item \textbf{GPUs:} The project has used a CUDA-enabled NVIDIA GTX 1080 GPU to accelerate the training of the CNN.
\end{itemize}
\subsubsection{Data}
Machine Learning and hence Deep Learning are both strongly connected to the field of data science. That's probably why you don't hear too little about Al being all about (Big) Data.
\subsubsection*{Tensors}\label{tensors}

\vspace{5mm}
\begin{displayquote} 
   \textit{All the data that is being processed, passed in and out of a neural network is represented using an n-dimensional array of numbers, called a tensor. }
\end{displayquote}
\vspace{5mm}

In Machine Learning, a tensor is used as a way to represent data that researchers and programmers use when dealing with neural networks. A tensor is the generalization of vectors and matrices and can be n-dimensional. A vector is just a one-dimensional or rank one tensor and a matrix a two-dimensional or rank two tensor. 
\vspace{5mm}

\begin{figure}[H]
   \centering
   \includegraphics[width=.45\textwidth]{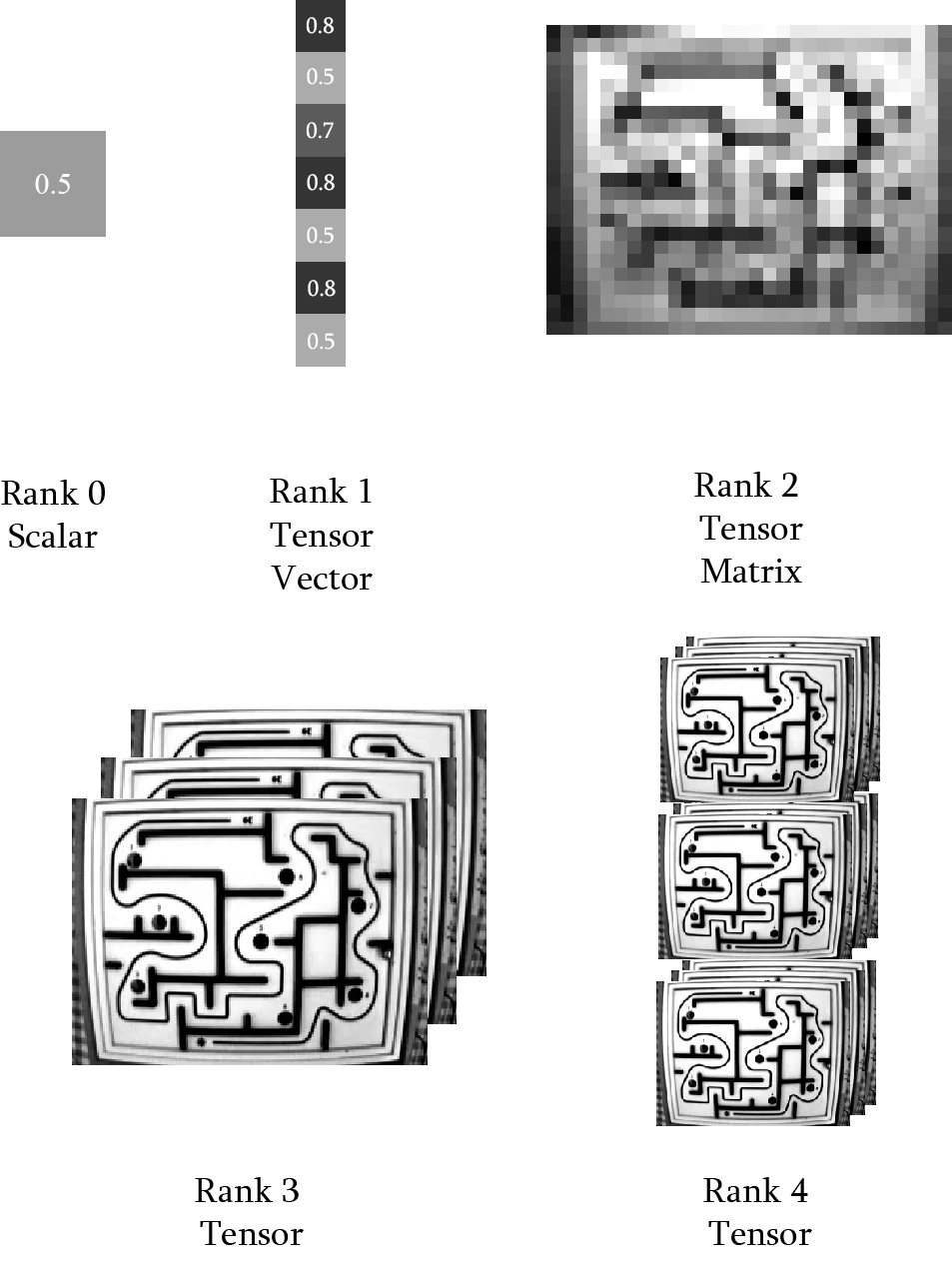}
   \caption[Source: Image created by authors of this paper.]{Graphical representation of a tensor with different ranks and dimensions.}
   \label{fig:TensorML}
\end{figure}
A tensor can be represented by an n-dimensional array of numbers. Its \textit{rank} is determined by the number of basis vectors needed to find a specific component. So if we had a two-dimensional array, we needed the row and the column of the component, so that tensor would have rank 2. The \textit{dimension} of a tensor is in the case of our array-representation simply determined by the number of rows or columns the array has. 
\begin{figure}[H]
   \centering
   \includegraphics[width=.3\textwidth]{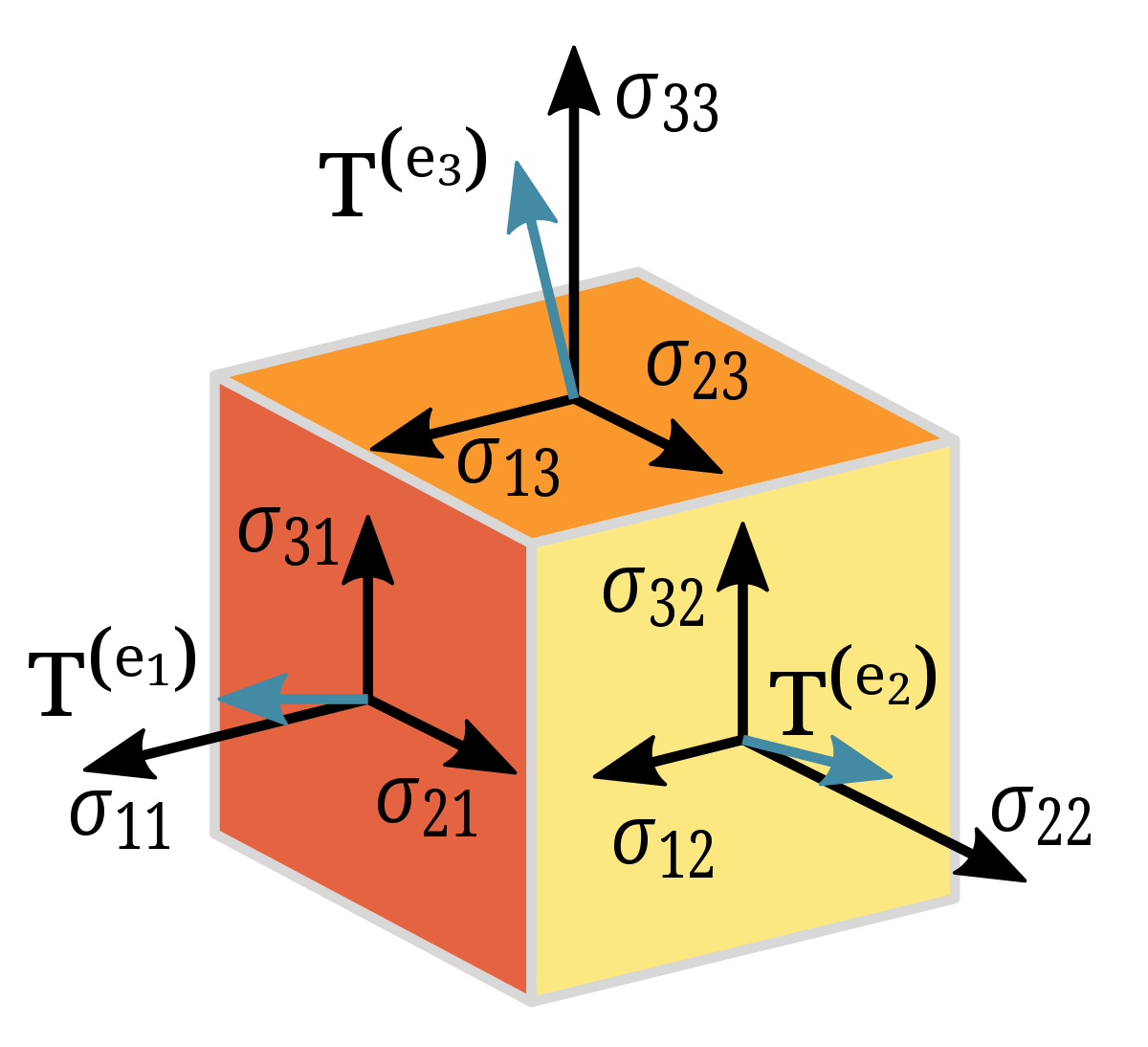}
   \caption[Source: https://en.wikipedia.org/wiki/Tensor]{Graphical representation of a tensor with dimension three and rank two (Also called the "Cauchy Stress Tensor").}
   \label{fig:tensor}
\end{figure}

The same observation (i.e. a ball rolling in a certain direction) can be represented by different coordinate systems. Different coordinate systems rely on different \textit{basis vectors}, so the \textit{components} in the mathematical objects used (i.e. a vector with a magnitude and a direction to represent the velocity of the ball) also have to be different to compensate for the change of the basis vectors to still represent the same observation \cite{tens2}. \\

The power of Tensors lies in the fact that they ensure that the observation they represent (in the form of a combination of a set of basis vectors and components) stays the same for every observer using any coordinate system \cite{tens}. This means that if you rotate your coordinate system, the ball keeps rolling "in the same direction" seen from an absolute standpoint because our way to represent an observation does not affect the actual physics going on.\\

To conclude for our work, Tensors have served as a way to represent the data that is passed through our neural network. That is when we first input the images from the camera, as well when the images are manipulated by the neural network when processing the data. It was always a Tensor, as you can see in figure \ref{fig:TensorML}. 

\subsubsection*{Dataset Creation}
When developing an AI algorithm, a substantial part of the work goes into preprocessing the data, so the neural network can learn properly. To create the training set for the CNN, we needed to go through a couple of steps. \\

After recording videos of the ball rolling around on the labyrinth (table tilted by hand), we had to label the video. We could do that by importing and activating the "DvsSliceAviWriter" filter in the jAER software. Once it is activated, there is a submenu called \textbf{"TargetLabeler"}, which lets you use your mouse to set the labels while watching the recorded video. Having done that, we used the jAER software to output a .avi video file with the recorded video, plus a .txt text file that contained the corresponding label locations/coordinates for every frame of the video that we had applied a label to using the mouse on the beforehand. To accelerate this process we made use of the program "After Effects", which already contains a tracker for different kinds of objects. With that, we could label our data faster and more accurately. In total, our dataset reached a size of about 30'000 frames. 

Since the video output of the DAVIS camera had the classical frames \textbf{(APS frames)} on the red channel and the event frames \textbf{(DVS frames)} on the green channel (blue is empty), the next step was to write a python script that could both extract all the frames from the .avi-formatted video plus separate the APS and DVS frames and save them into separate folders as .png images. With that, we have prepared the input to our CNN properly.\\

Since our project uses a "Supervised Learning" approach (see \ref{ML}), we have rate the output of the neural network automatically for the CNN to be able to continually improve its performance. This is where the labels come into play. The task now was to use the location information of the previously described .txt file to create so-called \textbf{heatmaps} which represent the ideal output we want from our neural network for each picture. By taking the coordinates of the .txt files and putting a Gaussian distribution around them, we could create the desired heatmaps. The purpose of these heatmaps is to help us tell the CNN how good or bad it is currently doing at telling us the location of the ball on the labyrinth, by providing the perfect solution. How we solved the task of rating the CNN as well as an example of a label is covered in \ref{lossfn} .\\

Once we had this collection of examples and their corresponding solutions prepared, we could start developing the architecture of the neural network itself. \\
\subsubsection*{Data Augmentation}
\vspace{5mm}
\begin{displayquote} 
   \textit{Data Augmentation is a technique used in Deep Learning to artificially increase the amount of data in the dataset, as well as to increase the performance and robustness of the neural network.}
\end{displayquote}
\vspace{5mm}

The technique's trick is to artificially increase the number of learning examples in the dataset by slightly modifying existing samples. With images you could, for example, rotate it, adjust the brightness, change the virtual lighting by shading the image, add fake highlight reflections, translate it, crop it, etc. as you can see below in figure \ref{fig:DA}. The three square images below have all been augmented. They can now all be used as a separate training example apart from the original image, thus multiplying the size of the dataset! You could of course implement more different forms of image augmentation, but transformations other than rotating, as well as adjusting the brightness and contrast were sufficient for good results of our study.\\

\begin{figure}[H]
   \centering
   \includegraphics[width=0.5\textwidth]{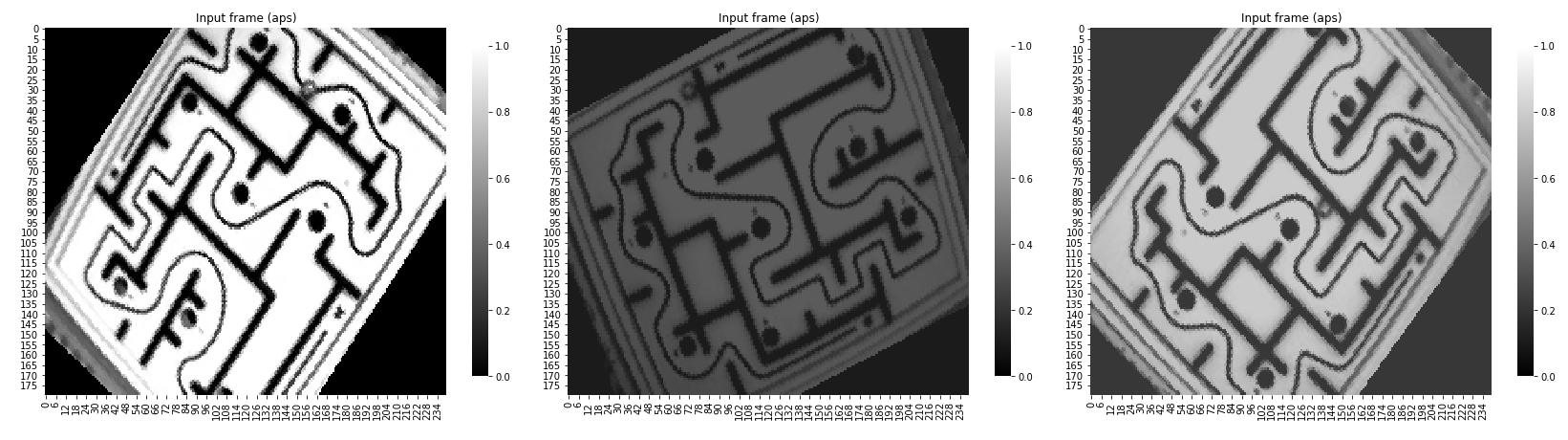}
   \caption[Source: Image created by authors of this paper.]{Examples of image augmentation used in our project.}
   \label{fig:DA}
\end{figure}

Our project used the technique of Data Augmentation with the intent to make the neural network more robust against different lighting conditions, slight rotations of the labyrinth etc. but also improve the CNNs accuracy by letting it train on a larger dataset. It also helped to increase the performance of the CNN under real-life conditions, because they have been simulated and trained on before.
\subsubsection{Network Architecture} \label{ANNs}
\subsubsection*{How does an ANN work?}

\vspace{5mm}
\begin{displayquote} 
   \textit{An Artificial Neural Network (ANN) is a form of AI, whose structure is strongly inspired by the human brain.}
\end{displayquote}
\vspace{5mm}
In this chapter, we will have a closer look at how a neural network works. After that, we present the architecture of the CNN we have developed for the practical project.\\

The first type of a neural network we will look at is the most simple form. It is called a "\textbf{Dense Neural Network (DNN)}" or "Fully-connected Neural Network". It consists of thousands to millions of nodes, that are organized into different layers. Each node is connected to all nodes in the previous layer as well as in the following layer. So there are even many more connections than there are nodes. This is why the individual layers are called "Dense Layers" or "Fully Connected Layers", the latter of which is often abbreviated with "FC" \cite{mitnn}.\\

\begin{figure}[H]
   \centering
   \includegraphics[width=.5\textwidth]{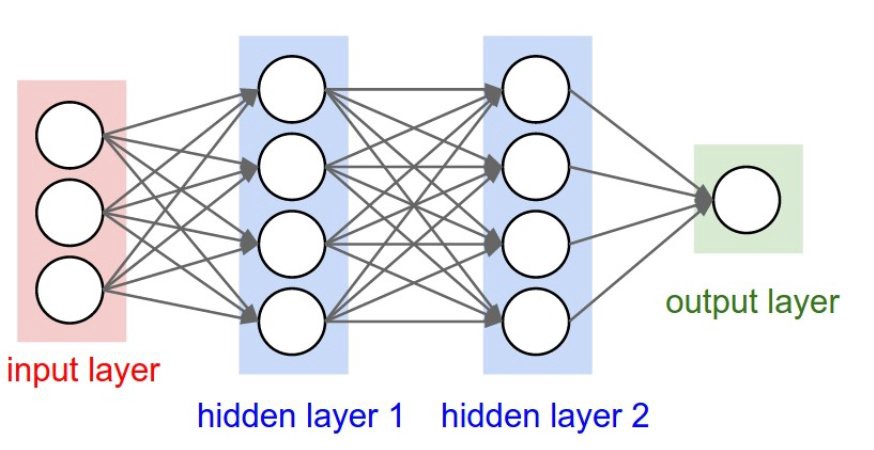}
   \caption[Source: https://towardsdatascience.com/building-a-deep-learning-model-using-keras-1548ca149d37]{Dense Neural Network.}
   \label{fig:dnn}
\end{figure}

Figure \ref{fig:dnn} shows a simplified DNN with fewer nodes and connections. The first layer of a neural network, in general, is called the "input layer" because this is where we feed in our data. The following layers are called "hidden layers" because we rarely access them directly and in most cases also don't exactly know what these layers represent logically. The last layer is the "output layer", this is what we receive back as a result of our neural network.

Let's have a look at the mathematics behind it.

\begin{figure}[H]
   \centering
   \minipage{0.3\textwidth}
     \includegraphics[width=\linewidth]{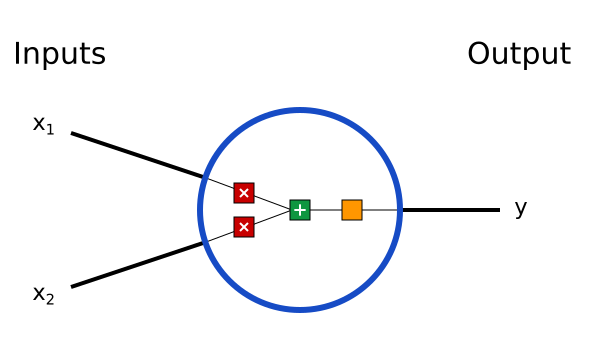}
     \caption[Source: https://towardsdatascience.com/machine-learning-for-beginners-an-introduction-to-neural-networks\linebreak-d49f22d238f9]{The structure of a neuron.}\label{fig:neuron}
   \endminipage\hfill
   \minipage{0.3\textwidth}
     \includegraphics[width=\linewidth]{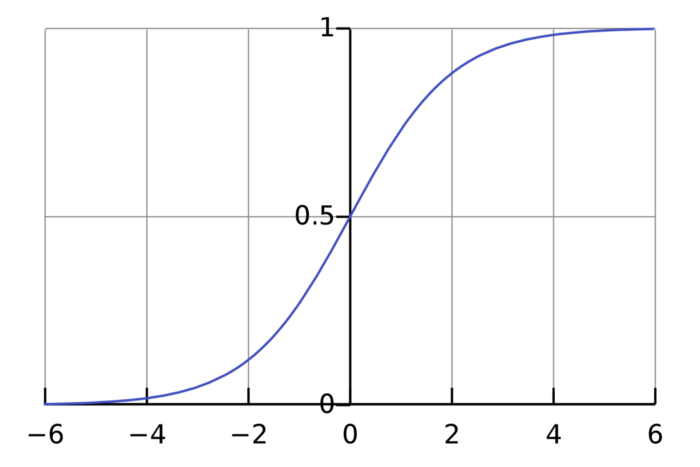}
     \caption[Source: https://towardsdatascience.com/machine-learning-for-beginners-an-introduction-to-neural-networks\linebreak-d49f22d238f9]{A Graph of the sigmoid activation function.}\label{fig:sigmoid}
   \endminipage\hfill
\end{figure}

A single neuron has the following structure (see figure \ref{fig:neuron}): It takes a set of inputs $\{x_1,\cdots, x_i, \cdots,x_n\}$, where n is the number of neurons it is connected to in the previous layer. In figure \ref{fig:neuron}, $n=2$. Each connection between two neurons has a \textbf{weight} $w_i$ associated with it. So before the inputs enter the neuron, they are multiplied by the weight that's associated with them.
$$x_i \rightarrow x_i*w_i$$
In the next step, we sum all of the weighted inputs and add a \textbf{bias} $b$ to it. The bias is a real number that helps to control how hard it is for a neuron to get excited (i.e. to output a large number to the next layer).
$$\sum_{i=1}^{n}(x_i*w_i) + b $$
As the last step, an \textbf{activation function} is used to bring our weighted sum into a predictable form. One function that is often used is called the "sigmoid function", denoted with the symbol $\sigma$. It compresses every input in the range between 0 and 1.
\begin{align*}
      \sigma(z) &= \frac{1}{1+e^{-z}} \\
      \lim_{z \to +\infty}\sigma (z) &= 1 \\
      \lim_{z \to -\infty}\sigma (z) &= 0 \\
\end{align*}
So finally, we can write the output $y$ of our neuron as:
$$y = \frac{1}{1+\exp{\left(-\sum_{i=1}^{n}(x_i*w_i)-b\right)}}$$
\newpage
For our network, the ReLU (Rectified Linear Unit) activation function
was used, as this is a commonly used activation function for CNNs which was first inspired by observations of real cortical neurons in the laboratory of the Institute of Neuroinformatics at UZH/ETH Zurich \cite{ReLUfirst}. The ReLU activation function is easy to compute and helps against the "vanishing gradient" problem, in which the gradient of the loss becomes
extremely small and thus the learning progress starts to stall (more on that in chapter \ref{tl}). The ReLU activation function sets every negative input to zero and leaves every positive input as it is.
It is mathematically expressed as:
$$ReLU(x) = max(0, x)$$
\begin{figure}[H]
 \centering
 \includegraphics[width=0.3\textwidth]{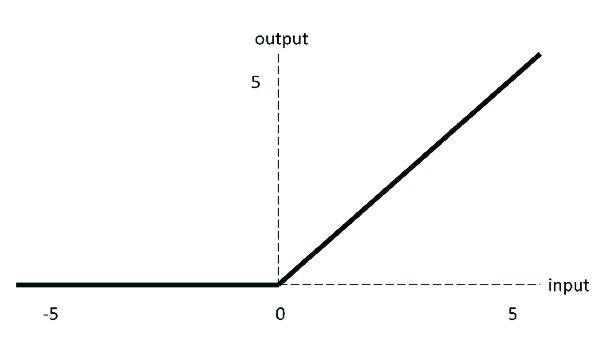}
 \caption[Source: Image created by authors of this paper.]{Graph of ReLU Activation function.}
\end{figure}

We now know how data gets fed through forward to produce our result. But how does a neural network learn? After outputting the result, the neural network gets feedback from a "Loss Function" which we will look at in chapter \ref{lossfn} . Once it knows how good or bad the produced result was, it uses two techniques called "Backpropagation" and "Gradient Descent" (See chapter \ref{tl}) to do better next time.
\subsubsection*{Convolutional Neural Networks}
A relatively newer form of a neural network, whose structure is inspired by the visual cortex of the human brain, is called a Convolutional Neural Network (CNN) or "ConvNet". As a consequence of their structural origin, the individual "neurons" in this type of neural net correspond to certain areas of the input image, known as the "Receptive Field" when talking about the visual system of humans. One of many differences to a DNN is that a CNN can capture the spatial information contained in an image fed into it. This fact makes CNNs excel at visual tasks such as object recognition/localization, which is why a CNN has also been used in the practical part of this paper. \\
\begin{figure*}[ht]
   \centering
   \includegraphics[width=\textwidth]{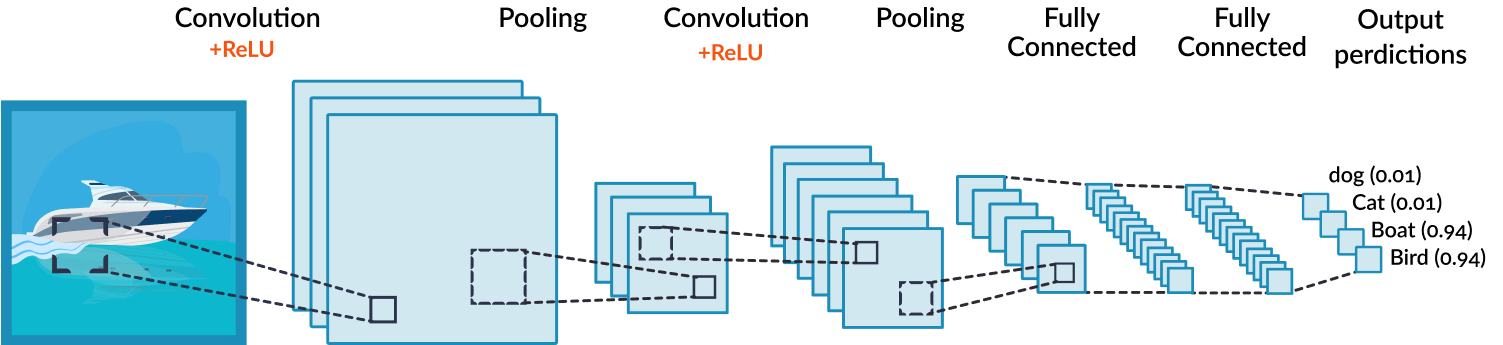}
   \caption[Source: https://missinglink.ai/guides/convolutional-neural-networks/convolutional-neural-network-tutorial\linebreak-basic-advanced/]{Example of a CNN.}
   \label{fig:cnn}
\end{figure*}

A typical CNN consists of two parts: First, we have a few "Convolutional Layers", whose job it is to extract the features of the input image and bring them into a form that is easier to process without losing the critical features. This information is then passed through a few "Fully Connected Layers", whose job it is to classify the received information into a category (See figure \ref{fig:cnn}) \cite{tdsconv}.\\

\begin{figure}[H]
   \centering
   \includegraphics[width=0.7\linewidth]{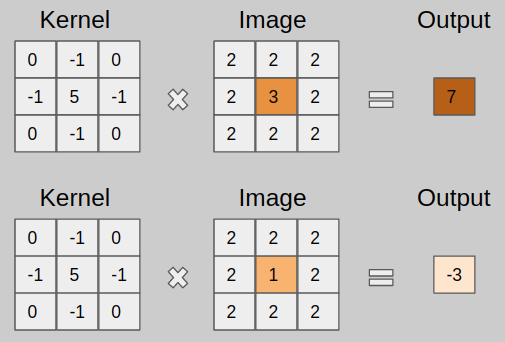}
   \caption[Source: https://towardsdatascience.com/types-of-convolution-kernels-simplified-f040cb307c37]{Example of a kernel.}
   \label{fig:conv}
\end{figure}

\begin{figure}[H]
   \centering
   \includegraphics[width=0.7\linewidth]{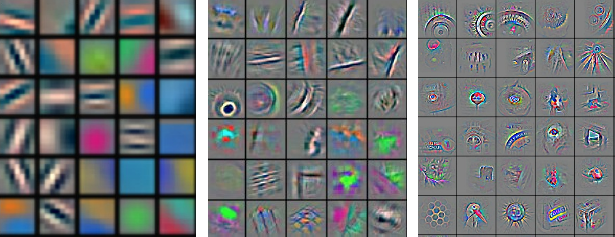}
   \caption[Source: https://stats.stackexchange.com/questions/362988/in-cnn-do-we-have-learn-kernel-values-at-every\linebreak-convolution-layer]{Different filter patterns.}
   \label{fig:enhance}
\end{figure}

A single "Convolutional Layer (ConvLayer)" consists of a set of \textbf{filters or kernels}, matrices that contain the "learnable" weights (The weights are what gets adjusted when the feature extractor learns). These filters are much smaller than the input image (See square on the image on the left side in figure \ref{fig:cnn}), so they propagate over the whole image, step by step, starting at the top left, ending at the bottom right of the image. After each step, the weights are multiplied with the underlying pixel-activations in that part of the image (See figure \ref{fig:conv}). The results of each step are then put together to form a new image. This output image is then used as input to the next ConvLayer. Depending on the filter weights, different aspects of the image are enhanced. In figure \ref{fig:enhance}, the patterns the individual filters of a CNN look for were visualized. In the left rectangle of figure \ref{fig:enhance}, we can see that edges at different angles and certain colors are being enhanced. Moving to the right, we can see the learned filters from following ConvLayers, which are a little more complex patterns, formed out of the simple edges and colours to the left. This goes on for the rest of the layers. The patterns the filters look for will change depending on what type of images the CNN was trained on.
In figure \ref{fig:cnn}, besides the ConvLayers and the Fully Connected Layers, we can also see so-called "\textbf{Pooling Layers}". Two types of pooling operations often used are called "average-pooling" and "max-pooling". Average-pooling outputs the average pixel-activation of the current set of pixels it is applied to. The max-pooling operation outputs the maximum pixel-activation out of the patch it is applied to. These layers help to reduce the size of the input, as well as the amount of redundant information.
\begin{figure}[t]
   \centering
   \includegraphics[width=.3\textwidth]{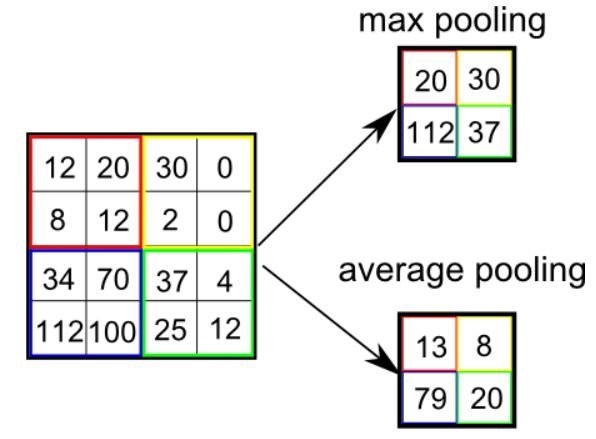}
   \caption[Source: https://www.quora.com/What-is-max-pooling-in-convolutional-neural-networks]{Average and Max Pooling operation on a 4x4 matrix.}
   \label{fig:pool}
\end{figure}

\subsubsection{Loss Function} \label{lossfn}
When training a neural network, you generally need to tell it in some way how it is doing in terms of delivering the output that you want it to. Here we can use a "Loss Function" (or also "Cost Function"). It is a function that takes two inputs: one is the output of the ANN, the other is a label that shows the desired output. It then in some way computes a "difference" between the two and outputs a number, that gets fed back into the ANN and is the origin for the learning process. The higher the number, the more different the output is from the desired output. This means that the lower this number (called "Loss" or "Cost"), the better the ANN is performing.\\

\begin{figure}[H]
   \centering
   \includegraphics[width=0.7\linewidth]{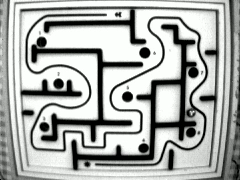}
   \caption[Source: Image created by authors of this paper.]{Example of a frame used for training.}\label{fig:frame}
\end{figure}

\begin{figure}[H]
   \centering
   \includegraphics[width=0.7\linewidth]{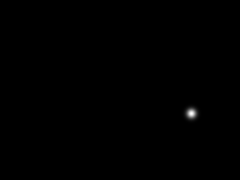}
   \caption[Source: Image created by authors of this paper.]{Example of a label used for training.}\label{fig:label}
\end{figure}

\begin{figure}[H]
   \centering
   \includegraphics[width=.4\textwidth]{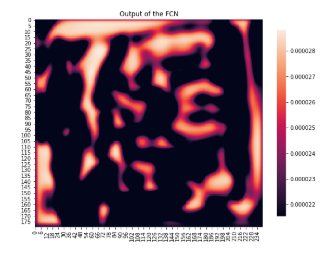}
   \caption[Source: Image created by authors of this paper.]{Output of our CNN at an early training stage.}
  \label{fig:earlytrainstage}
\end{figure}

\begin{figure*}[ht]
   \centering
   \includegraphics[width=1\textwidth]{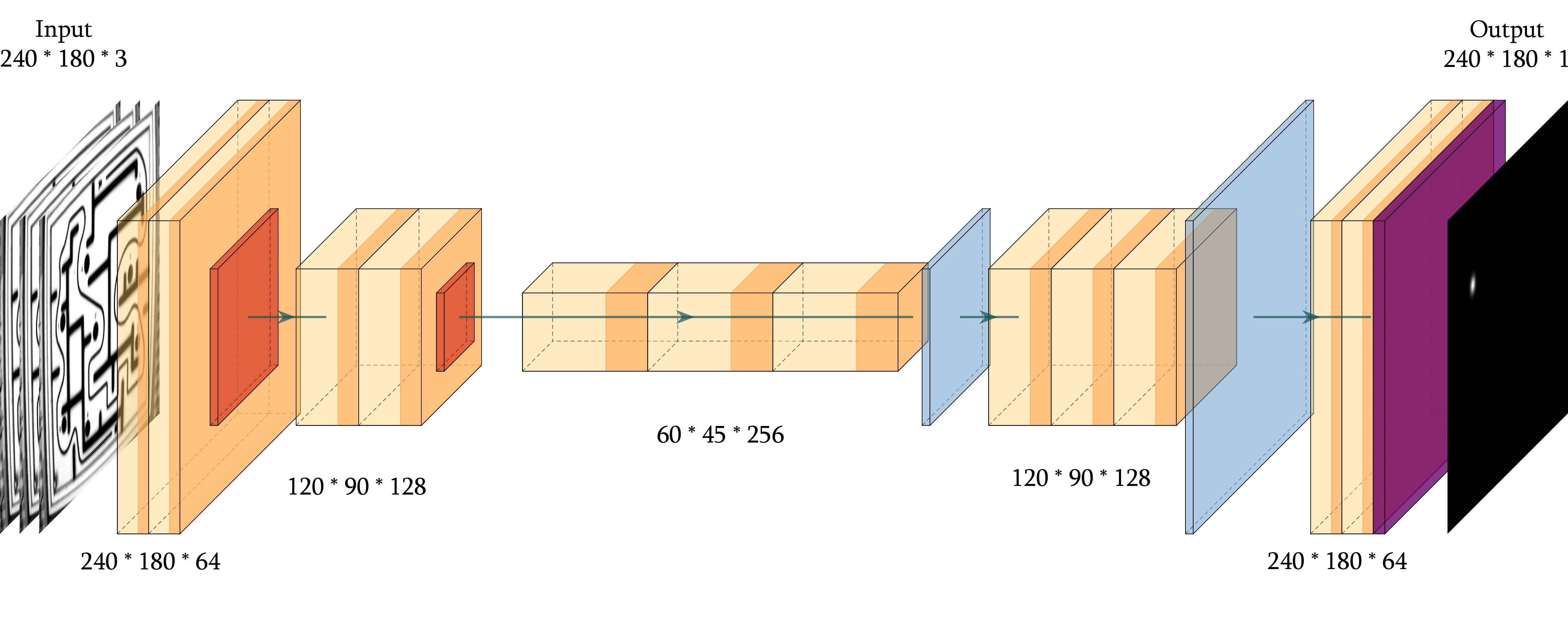}
   \caption[Source: Image created by authors of this paper.]{CueNet V2: \legendsquare{fill=ConvOrange}~Conv Layer, \legendsquare{fill=ReLUOrange}~ReLU, \legendsquare{fill=MaxRed}~MaxPooling Layer, \legendsquare{fill=UpBlue}~Upsampling Layer, \legendsquare{fill=SoftPurple}~Softmax Layer}
   \label{fig:CueNet_Arch}
\end{figure*}

The heatmaps we are using to represent our labels and network output are, again, just a 2D-matrix of numbers, in which every number stands for the activation of a pixel. In figure \ref{fig:frame}, there is a typical input image (that we got by extracting the red channel of the DAVIS video feed and converting it to black and white frames). In figure \ref{fig:label}, you can see a typical label picked out of the training dataset. These two figures are a input/label pair, which means that at the same coordinates of the white blob on the label, the cueball is at on the input image.
Now, at the beginning of training, the CNN has no clue what it should do and all the weights are initialised randomly. The output heatmap looks something like what you can see in figure \ref{fig:earlytrainstage}. Remember, the desired output would be something like in figure \ref{fig:label}. Also note that the activations are very small, as you can see on the scale to the right side of the heatmap. The reason for that is the architecture of our CNN, specifically the last layer, as you can see in the last part of figure \ref{fig:CueNet_Arch} . The whole heatmap can be looked at as a sort of "probability distribution" on where the CNN thinks the ball is.
Now we need a way to tell the network that this is not what we want. We do it by feeding back a big number to the neural network that stands for a big difference to our desired output. When dealing with images, the most simple and obvious way to calculate the difference between two pictures is to iterate over every pixel-position of the images and calculate the difference between the two pixel-values that are at the same location. You can then add up all those numbers or take the mean and you have the difference between those two pictures expressed in the form of a number.\\ 
           
The way just described is often used (also in this project) and it is called "\textbf{L1-Loss}". It can be expressed as the sum of absolute differences or mathematically as 
$$ Loss = \sum_{i=0}^{n} \ \lvert a_{L} - a_{O} \rvert$$
where the activation of a pixel in the label image is denoted by $a_{L}$, $a_{O}$ for the output image, respectively.

Modified versions of L1-Loss were put to the test during the development of CueNet (the CNN developed in this project), hoping it would learn faster, but none of the approaches was superior to the classical L1-Loss.

\subsubsection{Training the Network} \label{tl}
\subsubsection*{Gradient Descent}
After preprocessing the data, choosing the architecture of the network and declaring the loss function it is time to train it. In the beginning, all the weights and biases are randomly initialised. The network thus creates random outputs. With the loss function, you can measure how well the network is performing on a given input which alone doesn't get us anywhere. We have to find a way to tell the network how it should change its weights and biases so it can increase its performance and thus "learn".

To illustrate how this can be done, we will look at a simple Loss Function with just two inputs and one output expressed as: $$Loss = L(x, y)$$
If we plot the input on the x- and y-axis and the output on the z-axis we could get the following graph:
\begin{figure}[H]
   \centering
   \includegraphics[width=.4\textwidth]{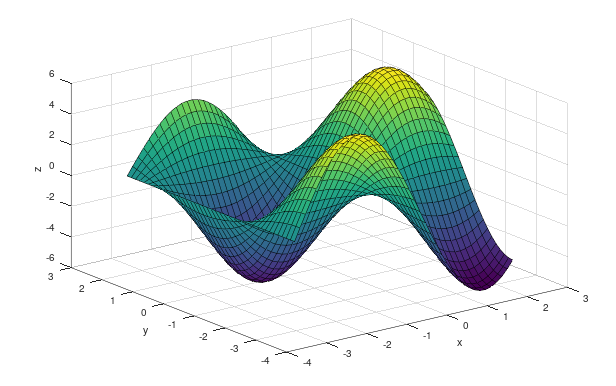}
   \caption[Source: https://mc.ai/gradient-descent-an-optimization-method-used-in-machine-learning/]{3D Plot of an example Loss Function}
   \label{fig:3Dloss}
\end{figure}

Our goal is to find the direction in this input space which decreases the loss most quickly, because the lower the loss, the better our network performs. The gradient of a function gives you a gradient vector whose direction points in the direction of steepest ascent. If we take the negative of the gradient vector, we get the direction of quickest descent. The gradient is calculated by packing all the partial derivative information of our loss function together.

\[
   \nabla L(x, y) =
\begin{bmatrix} \frac{\partial f}{\partial x} \\ \\\frac{\partial f}{\partial y} \end{bmatrix} 
\]

To find a local minimum we have to compute the gradient over and over while taking small steps in the inversed direction of the gradient vector. This process is called \textbf{Gradient Descent} and also applies to neural networks, many of which create input spaces with millions of dimensions. There are many ways to improve this process, such as "Stochastic Gradient Descent (SGD)", which we won't discuss here in detail.

\begin{figure}[H]
   \centering
   \includegraphics[width=.35\textwidth]{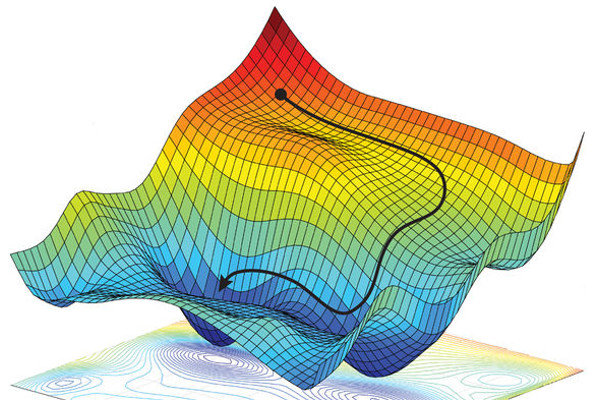}
   \caption[Source: https://blog.paperspace.com/part-3-generic-python-implementation-of-gradient-descent-for-nn\linebreak-optimization/]{Gradient Descent}
   \label{fig:GradientDescent}
\end{figure}

For us to take small steps on the plane the loss function needs to have a continuous output. This is the reason why artificial neurons have continuous ranging activations, unlike biological neurons which activations behave in a binary way.\\

Now that we know how to descend the valley of our loss function if we know the perfect direction, we need to figure out how to find that "direction" in the multi-million-dimensional input space of our loss function each step. This is where the Backpropagation algorithm comes into play.

\subsubsection*{Backpropagation}
\begin{figure}[H]
   \centering
   \includegraphics[width=.25\textwidth]{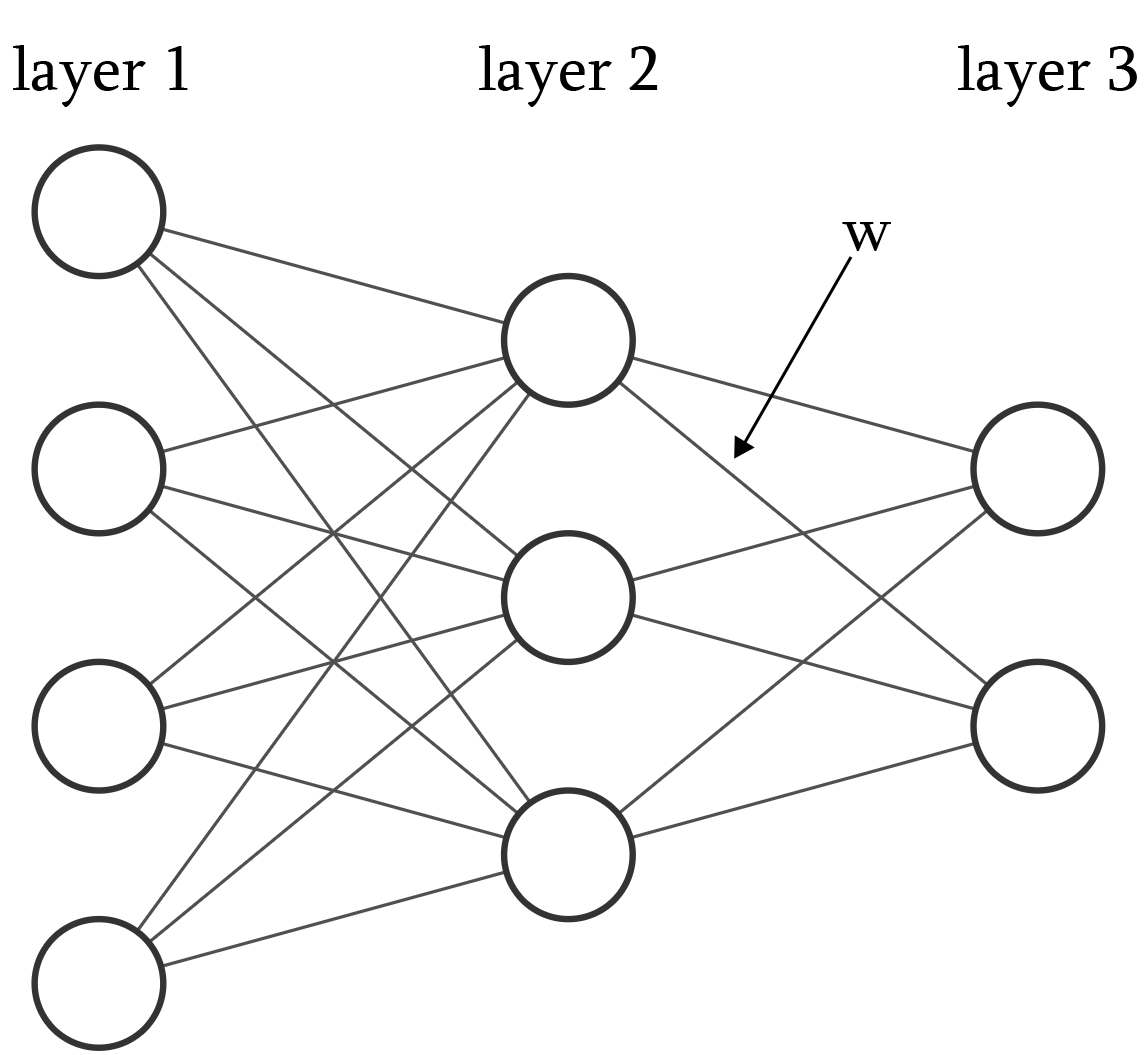}
   \caption[Source: Image created by authors of this paper.]{Example network to illustrate notation.}
   \label{fig:BPW}
\end{figure}

Calculating the gradient of a neural network is a lot harder as there are many more parameters we can change and their impact on the final loss function is dependent on changes to other parameters. The core of Backpropagation is an expression for the partial derivative  ${\partial L} / {\partial w}$ of the loss function $L$ with respect to any weight $w$ or bias $b$ in the network. This expression tells us how fast the loss changes when we shift the weights and biases. Going further we will rely on understanding the following notation which lets us refer to weights and biases in the network in an easy way. $w_{jk}^l$ will be used to denote the weight for the connection from the $k^{th}$ neuron in the $(l - 1)^{th}$ layer to the $j^{th}$ neuron in the $l^{th}$ layer. As an example, we will look at the indicated weight $w$ in figure \ref{fig:BPW}. The weight $w$ is the connection between the first neuron in the second layer to the second neuron in the third layer of the network. Therefore the weight $w$ is notated as $w_{21}^3$.
To refer to the network's biases and activations we use a similar notation. For the bias of the $j^{th}$ neuron in the $l^{th}$ layer, we use $b_j^l$. And for the activation of the $j^{th}$ neuron in the $l^{th}$ layer, we use $a_j^l$. With that the indicated bias $b$ and activation $a$ in the figure below are notated as $b_2^1$ and $a_2^3$.
\begin{figure}[H]
   \centering
   \includegraphics[width=.25\textwidth]{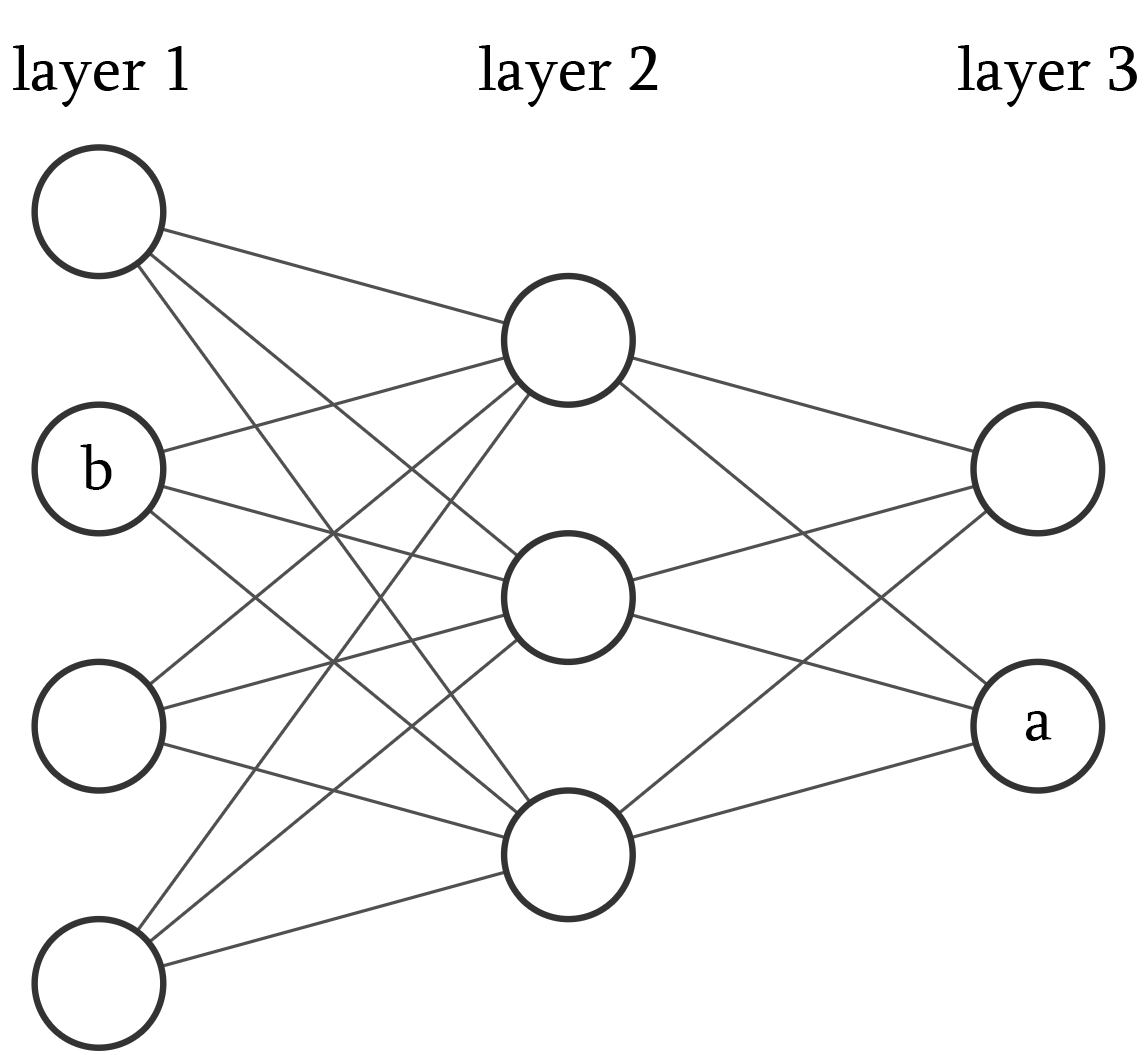}
   \caption[Source: Image created by authors of this paper.]{Example network to illustrate notation.}
   \label{fig:BPBA}
\end{figure}
With these notations, the activation $a_j^l$ is related to the activations in the $(l - 1)^{th}$ layer by the following equation: 
$$a_j^l = \sigma \Big(\sum_k w_{jk}^l a_k^{l-1} + b_j^l \Big)$$
where the sum is above all neurons $k$ in the $(l - 1)^{th}$ layer. The sigmoid activation function is just used as an example and can be exchanged with any activation function. 

We will now rewrite this expression in a matrix form to make our expression more readable and intuitively understandable. We define a weight matrix $w^l$ for each layer $l$ where the entries of the weight matrix are the weights connecting to the $l^{th}$ layer of neurons, that is, the entry in the $j^{th}$ row $k^{th}$ column is $w^l_{jk}$. Analogously the bias vector is defined for each layer $l$ by $b^l$ and the activation vector by $a^l$. The components of the bias and activation vectors are the values $b^l_j$ or $a^l_j$ for each neuron in the $l^{th}$ layer. Finally, we must vectorize our activation function $\sigma$. We do this by applying the function to every element in a vector $v$ and use the notation $\sigma(v)$ to denote this elementwise application of a function. We can now rewrite our last equation in a compact vectorized form
$$a^l = \sigma(w^l a^{l-1} + b^l).$$

Below is the written out activation vector of the $3^{rd}$ layer of the example network in figure \ref{fig:BPBA}.

$$ 
	\begin{bmatrix} 
		a_1^3 \\ 
		a_2^3  
   \end{bmatrix} 
   = \sigma \Bigg(\begin{bmatrix}
      w_{11}^3 & w_{12}^3 & w_{13}^3  \\
      w_{21}^3 & w_{22}^3 & w_{23}^3
   \end{bmatrix}
   \times
   \begin{bmatrix}
      a_1^2\\
      a_2^2\\
      a_3^2
   \end{bmatrix}
   + 
   \begin{bmatrix}
      b_1^3 \\
      b_2^3 
      
   \end{bmatrix}
   \Bigg)
$$

When we compute $a^l$ with our equation above we compute the intermediate quantity $z^l = w^l a^{l-1} + b^l$, which is the weighted input to the neurons in layer $l$, this $z^l$ will be useful later. With this, our equation above can be rewritten as $a^l = \sigma(z^l)$.

To use Backpropagation we must make two main assumptions about the shape of our loss function. To state these assumptions we will look at the quadratic lost function as an example, which has the form 
$$L = \frac{1}{2n} \sum_x \lVert y(x) - a^L(x)\lVert^2.$$

where $n$ is the total amount of training samples. The sum is over all the samples in the training set, $y(x)$ is the corresponding desired output. $L$ is the number of layers in the network and $a^L(x)$ is the vector of the output activations from the network for the given input $x$.

The first assumption we have to make about our loss function $L$ is that the loss function can be written as an average $L = \frac{1}{n} \sum_x L_x$ over loss functions $L_x$ for the individual training examples $x$. For our loss function, this is the case, as the loss for a single training example equals $L_x = \frac{1}{2} \lVert y - a^L \lVert^2$.

The reason for this is that we try to approximate the partial derivatives $\delta L/ \delta w$ and $\delta L/ \delta b$ by calculating the partial derivatives $\delta L_x/ \delta w$ and $\delta L_x/ \delta b$ for each training example $x$ and average it. With that in mind, we will remove the subscript $x$ and write the loss $L_x$ as $L$.

The second assumption about our loss function is that it can be rewritten as a function that takes the output of our network as the input. This applies to our loss function $L$ as it can be written for a single training example $x$ as 

$$L = \frac{1}{2} \lVert y - a^L \lVert^2 = \frac{1}{2} \sum_j (y_j - a^L_j)^2$$

and by that is a function of the output activations of our network. One may wonder why we do not take the loss $L$ as a function of the desired output $y$ or the training example $x$. This is because $x$ and $y$ are fixed parameters. 

We will now introduce the intermediate quantity $\delta^l_j$, which we will call the $error$ in the $j^{th}$ neuron in the $l^{th}$ layer to ultimately compute $\delta L / \delta w^l_{jk}$ and $\delta L / \delta b^l_{jk}$. Backpropagation will enable us to compute this error $\delta^l_j$ and then will relate $\delta^l_j$ to $\delta L / \delta w^l_{jk}$ and $\delta L / \delta b^l_{jk}$. To elucidate this, imagine an error which lies in the $j^{th}$ neuron in the layer $l$. When the input enters the neuron, the error adds a little change $\Delta z^l_j$ to the neuron's weighted input, which changes the neuron's output $\sigma (z^l_j)$ to $\sigma (z^l_j + \Delta z^l_j)$. We must now find a $\Delta z^l_j$ that minimizes the loss. If $\frac{\delta L}{\delta z^l_j}$ is large then we can lower the loss a lot by choosing $\Delta z^l_j$ to have the opposite sign to $\frac{\delta L}{\delta z^l_j}$. Reciprocally if $\frac{\delta L}{\delta z^l_j}$ is small making changes to $\Delta z^l_j$ won't alter the loss by much. As a consequence, we can already tell which neuron is near the desired state. We define this error $\delta^l_j$ in the $j^{th}$ neuron and $l^{th}$ layer by 

$$
\delta ^l_j = \frac{\delta L}{\delta z^l_j}.
$$

As before we will use $\delta^l$ as the vector of errors in layer $l$ and equals 
\begin{equation}
\delta^L_j = \frac{\delta L}{\delta a^L_j} \sigma'(z^L_j).
\tag{1}
\end{equation}
for the output layer.

This intuitively makes sense as the first term $\delta L / \delta a^L_j$ shows us how fast the loss is changing as a function of the $j^{th}$ output activation. If $L$ doesn't depend much on the given output by the $j^{th}$ neuron then $\delta^L_j$ will be small. The second term $\sigma'(z^L_j)$ shows us how fast the activation function $\sigma$ changes at 
$z^L_j$. 

Equation 1 is easily computed as $z^L_j$ will already be computed when passing the training data through the network and it's only a small addition to compute $\sigma'(z^L_j)$. The form of $\delta L / \delta a ^L_j$ will be dependent on the form of our loss function. For example, when using the already known quadratic loss function $L = \frac{1}{2} \sum_j(y_j - a^L_j)^2$ it would be $\delta L/\delta a^L_j = (a^L_j - y_j)$ which isn't hard to compute.

While equation 1 is a componentwise expression for $\delta^L$ we want a matrix-based form which we get by rewriting the equation as

\begin{equation}
\delta^L = \nabla_a L \odot \sigma'(z^L).
\tag{2}
\end{equation}

$\nabla_aL$ is a vector whose components are the partial derivatives $\delta L / \delta a^L_j$ and thus the rate of change of $L$ with respect to the output neurons. We will use these equations interchangeably. Our example quadratic loss function equals $\nabla_aL = (a^L - y)$ and the fully matrix-based form can be written as 

$$\delta^L = (a^L-y) \odot \sigma'(z^L).$$

The next equation we will look at will be

\begin{equation}
   \delta^l = ((w^{l+1})^T \delta^{l+1}) \odot \sigma'(z^l). 
\tag{3}
\end{equation}

With this equation, we can calculate the error $\delta^l$ in relation to the error of the next layer $\delta^{l + 1}$. By applying the transpose weight matrix $(w^{l + 1})^T$ we are moving the error back through the network. By taking the Hadamard product $\odot \sigma'(z^l)$ we move the error backwards through the activation function, leaving behind the error $\delta^l$ in the weighted input in layer $l$. We are propagating backwards through the network which is the reason it is called Backpropagation. 

Now with the combination of equation 2 and 3, we can compute the error $\delta^l$ for any layer in the network. We start by calculating $\delta^L$ with the $2^{nd}$ equation, then use the $3^{rd}$ equation to compute $\delta^{l -1}$ over and over for all remaining layers of the network.

The next equation we will introduce lets us calculate the rate of change of the loss with respect to any bias in the network as we know how to calculate $\delta^l_j$. 

\begin{equation}
   \frac{\partial L}{\partial b^l_j} =
  \delta^l_j.
  \tag{4}
\end{equation}

Equation 5 will enable us to calculate the rate of change of the loss with respect to any weight in the network.

\begin{equation}
     \frac{\partial L}{\partial w^l_{jk}} = a^{l-1}_k \delta^l_j.
     \tag{5}
\end{equation}

As we know how we can calculate the terms of the quantities $\delta^l$ and $a^{l-1}$ we can now compute equation 5.

The Backpropagation algorithm will go through five steps, the first being setting the corresponding activation $a^1$ for the input layer. The next step is to feedforward the input activation. For each $l - L$ we compute $z^l = w^l a^{l-1} + b^l$ and $a^l = \sigma(z^l)$. The third step is to compute the error vector for the last layer $L$ with $\delta^{L}
= \nabla_a L \odot \sigma'(z^L)$. The fourth step is to backpropagate the error back through the network by computing $\delta^{l} = ((w^{l+1})^T \delta^{l+1}) \odot
\sigma'(z^{l})$ for all the layers. The last step is to calculate the gradient of the loss function with $\frac{\partial L}{\partial w^l_{jk}} = a^{l-1}_k \delta^l_j$, $\frac{\partial L}{\partial b^l_j} = \delta^l_j$ and change the weights and biases accordingly. 

As we have already discussed we calculate the gradient for one single training example a time. This is because calculating the gradient for all the examples would take too much memory and therefore increase the computational cost. We combine Backpropagation with a learning algorithm such as Stochastic Gradient Descent, in which we compute the gradient for a chosen batch-size $m$ and take a learning step based on the gradient of this batch, by doing this we decrease the accuracy but can learn a lot faster. \cite{backpropagationvid}\cite{backpropagationweb}

\subsubsection*{Training phases}
\begin{figure}[H]
   \centering
   \includegraphics[width=.4\textwidth]{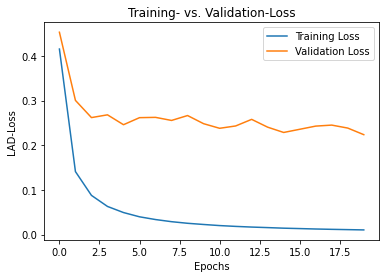}
   \caption[Source: Image created by authors of this paper.]{Training Loss vs Validation Loss.}
   \label{fig:trainlossvsvalidationloss}
\end{figure}

In our code, we define two sections: the training loop and the validation loop. In the first step, we go through the training loop where we calculate the partial derivatives for each training example and update the weights and biases according to the calculated gradient which gets multiplied by a \textit{learning rate} $\alpha$ which lets us adjust the steps the network takes so that we don't jump over local minima. After we went through all the training data we jump to the validation loop in which we calculate the loss for each example in the validation set and average it. This time we won't change the weights and biases as we want to check if the network can generalize what it has learned from the training set. These two phases will be repeated multiple times in so-called \textit{epochs}, which is a pass over all the training data. The loop won't stop until the validation loss does not decrease anymore. At this point, the network starts to overfit, which means that it has started to simply memorize the data in the training set and doesn't generalize well. In figure \ref{fig:trainlossvsvalidationloss} you can see how both the training loss and validation loss behaved when going through 20 epochs in our experiment.

\subsection{Analysis}

CueNet consists of an abridged version of the VGG16 convolutional neural network with a subsequent deconvolutional neural network. The 240 x 180-pixel input with a depth of 3 is passed through the VGG16 part, which reaches a depth of 256 and a resolution of 60 x 45 in the middle of the CueNet. In the DeconvNet part, upscaling layers are used to recover the dimensionality loss from the max-pooling layers. Due to the Softmax layer as the last layer of the network, the output produces a probability heatmap for the ball's location where each pixel carries a probability for the ball being there. All these probabilities sum up to 1, as the ball has to be somewhere on the board. The coordinates of the cueball equal the coordinates of the activation-peak on the heatmap.
Our input has a depth of 3 channels because we want to give 3 consecutive images to CueNet, hoping that it can learn something about the behaviour of a ball, e.g. a ball cannot jump from one side of the labyrinth to the other in consecutive images. For comparison purposes, we have created an identical version of the CueNet, which can only use one image as an input. \cite{TrackNet}

\begin{figure}[H]
   \centering
   \includegraphics[width=.4\textwidth]{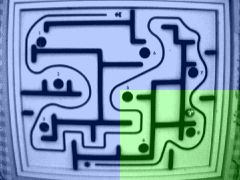}
   \caption[Source: Image created by authors of this paper.]{\legendsquare{fill=blue}~Training Set, \legendsquare{fill=green}~Validation Set.}
   \label{fig:trainvalidlocation}
\end{figure}

Our testing set is used to evaluate the performance of the CueNet, distinguishing between CueNet V1, which takes one image as an input and CueNet V2, which takes three consecutive images as an input. The rest of the dataset is separated into a training set and a validation set, depending on the true position of the ball (see figure \ref{fig:trainvalidlocation}).

\begin{table}[H]
   \resizebox{0.49\textwidth}{!}{%
   \begin{tabular}{|c|c|c|c|c|c|}
   \hline
   \multicolumn{1}{|c|}{Layer} & \multicolumn{1}{c|}{Filter Size} & \multicolumn{1}{c|}{Depth} & \multicolumn{1}{c|}{Padding} & \multicolumn{1}{c|}{Stride} & \multicolumn{1}{c|}{Activation Fn} \\ \hline
   Conv1                       & 3 x 3                            & 64                         & 1 x 1                        & 1                           & ReLU + BN                          \\ \hline
   Conv2                       & 3 x 3                            & 64                         & 1 x 1                        & 1                           & ReLU + BN                          \\ \hline
   Pool1                       & \multicolumn{5}{c|}{2 x 2 max pooling, Stride = 2, Padding = 0}                                                                                                 \\ \hline
   Conv3                       & 3 x 3                            & 128                        & 1 x 1                        & 1                           & ReLU + BN                          \\ \hline
   Conv4                       & 3 x 3                            & 128                        & 1 x 1                        & 1                           & ReLU + BN                          \\ \hline
   Pool2                       & \multicolumn{5}{c|}{2 x 2 max pooling, Stride = 2, Padding = 0}                                                                                                 \\ \hline
   Conv5                       & 3 x 3                            & 256                        & 1 x 1                        & 1                           & ReLU + BN                          \\ \hline
   Conv6                       & 3 x 3                            & 256                        & 1 x 1                        & 1                           & ReLU + BN                          \\ \hline
   Conv7                       & 3 x 3                            & 256                        & 1 x 1                        & 1                           & ReLU + BN                          \\ \hline
   UpS1                        & \multicolumn{5}{c|}{2 x 2 upsampling}                                                                                                                           \\ \hline
   Conv8                       & 3 x 3                            & 128                        & 1 x 1                        & 1                           & ReLU + BN                          \\ \hline
   Conv9                       & 3 x 3                            & 128                        & 1 x 1                        & 1                           & ReLU + BN                          \\ \hline
   Conv10                       & 3 x 3                            & 128                        & 1 x 1                        & 1                           & ReLU + BN                          \\ \hline
   \multicolumn{1}{|c|}{UpS2}  & \multicolumn{5}{c|}{2 x 2 upsampling}                                                                                                                           \\ \hline
   Conv11                      & 3 x 3                            & 64                         & 1 x 1                        & 1                           & ReLU + BN                          \\ \hline
   Conv12                      & 3 x 3                            & 64                         & 1 x 1                        & 1                           & ReLU + BN                          \\ \hline
   Conv13                      & 3 x 3                            & 64                         & 1 x 1                        & 1                           & ReLU + BN                          \\ \hline
   \multicolumn{6}{|c|}{Softmax}                                                                                                                                                                 \\ \hline
   \end{tabular}}
   \caption[Source: Image created by authors of this paper.]{\label{tab:CueNet}Parameters of CueNet.}
\end{table}

This is to ensure that CueNet generalizes and does not overfit. By separating the training data and validation data by the true location of the ball we make sure that there are not any nearly identical images in the training and validation set. If you were to split your sets by randomly shuffling your data or by time, there could be frames in the training and validation that are almost identical, which leads to the network not generalizing but rather memorizing the frames. To optimize the weight and biases, the Adam optimizer was used. We set the number of epochs to 20 to ensure that we find the least validation loss. The entire training using a batch size of 1 took about 8 hours.
CueNet is trained to produce the "correct" heatmap output. But to allow us to quantify the accuracy of CueNet, we define a tolerance range in which the output can lie and still be classified as correct. Since the ball is about 8 pixels wide, the tolerance range is 4 pixels around the center of the real position of the ball. The peak of the heatmap from the output of CueNet has to be in this area to be counted as a correct classification. The Positioning Error (PE) is the distance between both centers in pixels. The distribution of the PE is shown in figure \ref{fig:pedistribution}. 
\begin{figure}[H]
   \centering
   \includegraphics[width=.4\textwidth]{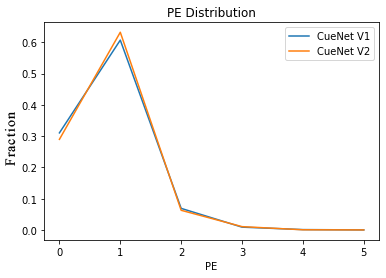}
   \caption[Source: Image created by authors of this paper.]{PE distribution of the two versions of CueNet.}
   \label{fig:pedistribution}
\end{figure} 
The x-axis of the plot shows us the PE and the y-axis the fraction of the occurrences for this PE. This distribution shows that 99.6\% of the classifications of CueNet V1 are in our set tolerance range of PE $<=$ 4 and 99.8\% of the classifications of CueNet V2 respectively.

Apart from the standard testing set, we have created a special dataset called "Heavy Test" to push CueNet to its limits. The recordings of "Heavy Test" were made in very poor lighting conditions, as a lamp was rotated around the labyrinth at a steep angle. This resulted in very strong and long shadows on the labyrinth, which should lead to a more difficult classification task. In addition to this, the view of the board was alternately covered with hands without covering the ball, which should show that CueNet can ignore distractions. 

\begin{figure}[H]
   \centering
   \includegraphics[width=.4\textwidth]{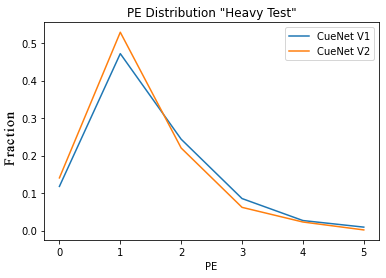}
   \caption[Source: Image created by authors of this paper.]{PE distribution of the two versions of CueNet on a more difficult testing set.}
   \label{fig:pedistribution}
\end{figure} 

In this benchmark CueNet V1 achieved 92.9\% accuracy with our given tolerance range of PE $<=$ 4 and CueNet V2 achieved an accuracy of 96.4\%. When comparing the mistakes both networks made you can see that CueNet V1 often loses track of the ball completely and thus labels the ball as somewhere completely else. By contrast, CueNet V2 locates the ball just barely outside of the tolerance range. We think that CueNet V2 is more robust because it can rely on the two older images and can therefore approximate the position of the ball with the learned trajectory patterns. In addition, it can compare changes between the consecutive images, making it easier to locate the ball since the ball is the fastest moving object. The potential tradeoff for CueNetV2's higher accuracy and robustness is higher computational cost, which could be important for the robotics control. 

As the last test, we combined some of the footage of our "Heavy Test" set to test the multi-object tracking capabilities of our network. Despite never having trained this, both versions of CueNet can track two objects simultaneously with sufficient result. If we had trained this we are confident that simultaneous tracking would not have been a problem since the last layer is a softmax layer and can easily split the probability distribution and thus track multiple objects.  
\begin{figure}[H]
   \centering
   \includegraphics[width=.4\textwidth]{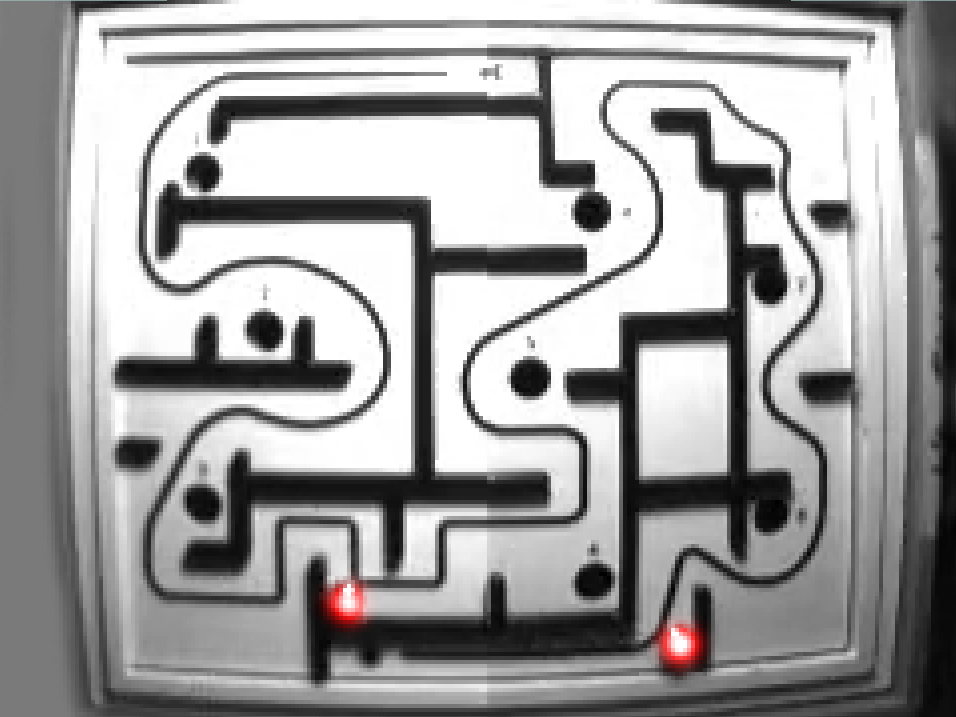}
   \caption[Source: Image created by authors of this paper.]{\legendsquare{fill=red}~Output of CueNet tracking two balls simultaneously, where the heatmap output in red is overlaid over the APS image.}
   \label{fig:twoballs}
\end{figure}

\paragraph{Conclusion} ~\\
In the context of this work, we developed a Fully Convolutional Neural Network named CueNet that can effectively detect and track the cueball on the used labyrinth. We also found that CueNet V2 outperforms CueNet V1, which indicates that using consecutive images as an input to the CNN increases its accuracy overall and especially in difficult situations. Additionally, both versions of CueNet could generalize well and thus were able to track multiple objects simultaneously, despite never having been trained on tracking more than one object. By extensively testing CueNet in difficult conditions, we could confirm the robustness/reliability of the algorithm under real-life conditions.

\section{Discussion}
\subsection{Applications of AI} \label{app}
In recent years, AI has found many new applications in a wide range of disciplines. In this section, we will cover some in which AI has made significant advances. 

\subsubsection*{Medicine}
Through the vast amount of generated and stored data in healthcare, AI today has found use in diagnosis, drug development and the creation of personalized treatments.

Deep Learning algorithms assist doctors in the evaluation of CT scans, MRI images, skin images etc. for the diagnosis of all kinds of diseases. Some of the algorithms can achieve accuracy of up to 99\% which is almost as good as a professional doctor \cite{medicalaccuracy}. The immense benefit of these algorithms is their low cost and speed. They can draw conclusions in seconds and the results can be shared across the world at the speed of light. This means that even people in poorer regions can receive a professional diagnosis for less money. \cite{medicaldeeplearning}

When designing the treatment for a patient there are many variables one has to take into account. Small differences in drug doses and treatment schedules can lead to different reactions from different patients. With the help of machine learning, doctors can find correlations between the patient's characteristics and the patient's response to a particular treatment. This helps with choosing the right treatment and thus maximizing the chance of the patient's survival. \cite{medicaldeeplearning}
\subsubsection*{Finance}

The use of AI in banking dates back to 1987 where the Security Pacific National Bank in the US started using an expert system for the detection of unauthorized use of debit cards \cite{aibanking}. Today machine learning is used to reduce the chance of fraud and financial crimes by monitoring behavioural patterns for anomalies and flag them for human review \cite{fraudprevention}.

In the world of trading, there is the never-ending competition of coming up with the best mathematical model to predict the future of the stock market. High-frequency trading represents one of the fastest-growing sectors in financial trading and of course, AI was found to be useful with this kind of problem. By 2010, 60\% of all trades were executed by computers of which most work with "simple" algorithms (Algorithmic Trading). Those algorithms can make tens of thousands of trades per second. Machine Learning, on the other hand, working a lot slower, can learn sophisticated patterns from past data that could indicate the next financial crisis, a bullish/bearish trend etc. Many banks, funds and trading firms have their entire portfolios managed purely by AI systems \cite{bankstrading}. There are ETFs (Exchange Traded Funds) whose portfolios are built with the help AI like ARK's Industrial Innovation ETF. Starting this year it gained 69.60\% compared to the average yearly 10\% that the S\&P 500 gains, which is quite an accomplishment. \cite{arkinvest}.

\subsubsection*{Transportation}

\begin{figure*}[ht]
   \centering
   \includegraphics[width=\textwidth]{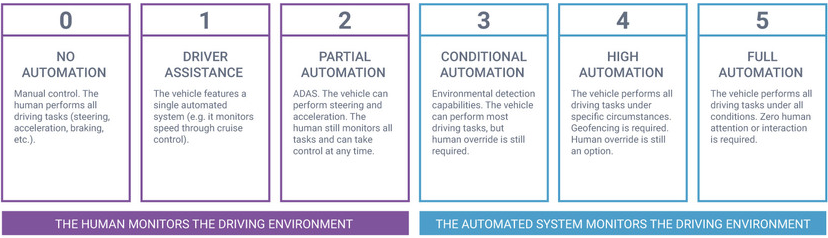}
   \caption[Source: https://www.synopsys.com/automotive/autonomous-driving-levels.html (modified)]{Levels of Vehicle Autonomy. }
   \label{fig:6levels}
\end{figure*}

We are on the verge of a revolution in the automotive industry, there are currently over 40 companies developing self-driving cars using AI. Tesla and even tech giants like Apple and Google are all developing their own systems \cite{40carcompanies}, of which Tesla's is the most advanced so far. The advantage of Tesla lies in its possession of a huge amount of real-life data. Their entire fleet collects data while driving and sends it back to Tesla, where they can train their neural network with it. Driving automation is subdivided into six levels (See figure \ref{fig:6levels}), of which we have only reached level 2 on commercial products up to this day.

Tesla has already demonstrated a level 5 system, but mainstream production of anything higher than level 2 has yet to come. The reason for this is partially the lack of technological capability but also the ethical and political questions that go along with autonomous driving. Another problem of autonomous systems is that they can be exploited and thus serious accidents could occur. In chapter \ref{reliability} we will talk more about the reliability and potential safety exploits of AI.

\subsubsection*{Computer Vision}

\begin{figure}[H]
   \centering
   \includegraphics[width=.4\textwidth]{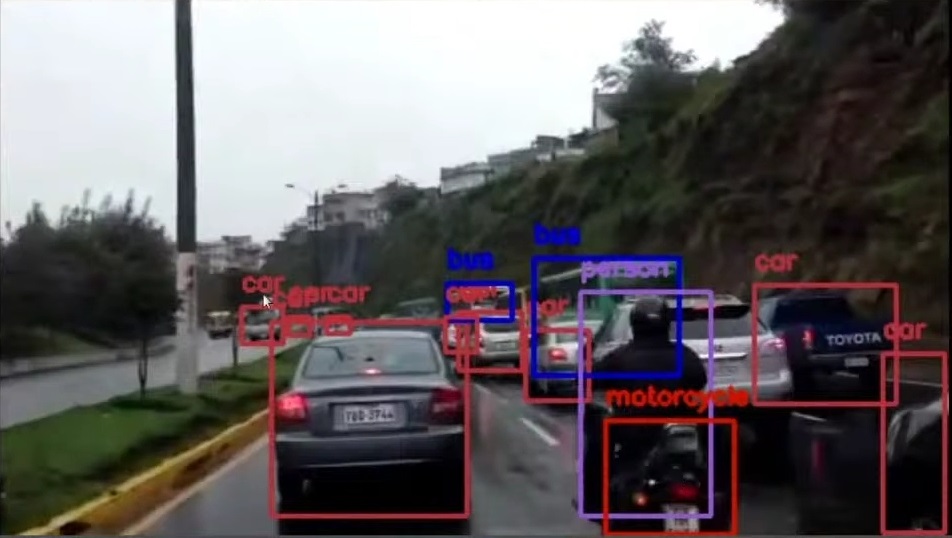}
   \caption[Source: https://www.wikiwand.com/en/Computer\_vision]{Tracking of vehicle movement using Artificial Intelligence.}
   \label{fig:cvvehicles}
\end{figure}

Image restoration, object recognition, video tracking, scene reconstruction, motion estimation are all sub-domains of computer vision \cite{cv}. The practical part of this paper goes into object recognition and video tracking, as we are trying to track a cueball from a continuous stream of frames. Object tracking and recognition are crucial for autonomous systems because they are built upon them. For them to be able to make decisions, they first have to see. In figure \ref{fig:cvvehicles}, an algorithm tracks and recognizes different kinds of vehicles.  

In image restoration, the goal is to remove unwanted noise from pictures or add missing details. This is typically done by feeding the network with millions of images until it can fill the missing content considering the other content of the image. \cite{imagerestoration} 

\begin{figure}[H]
   \centering
   \includegraphics[width=.4\textwidth]{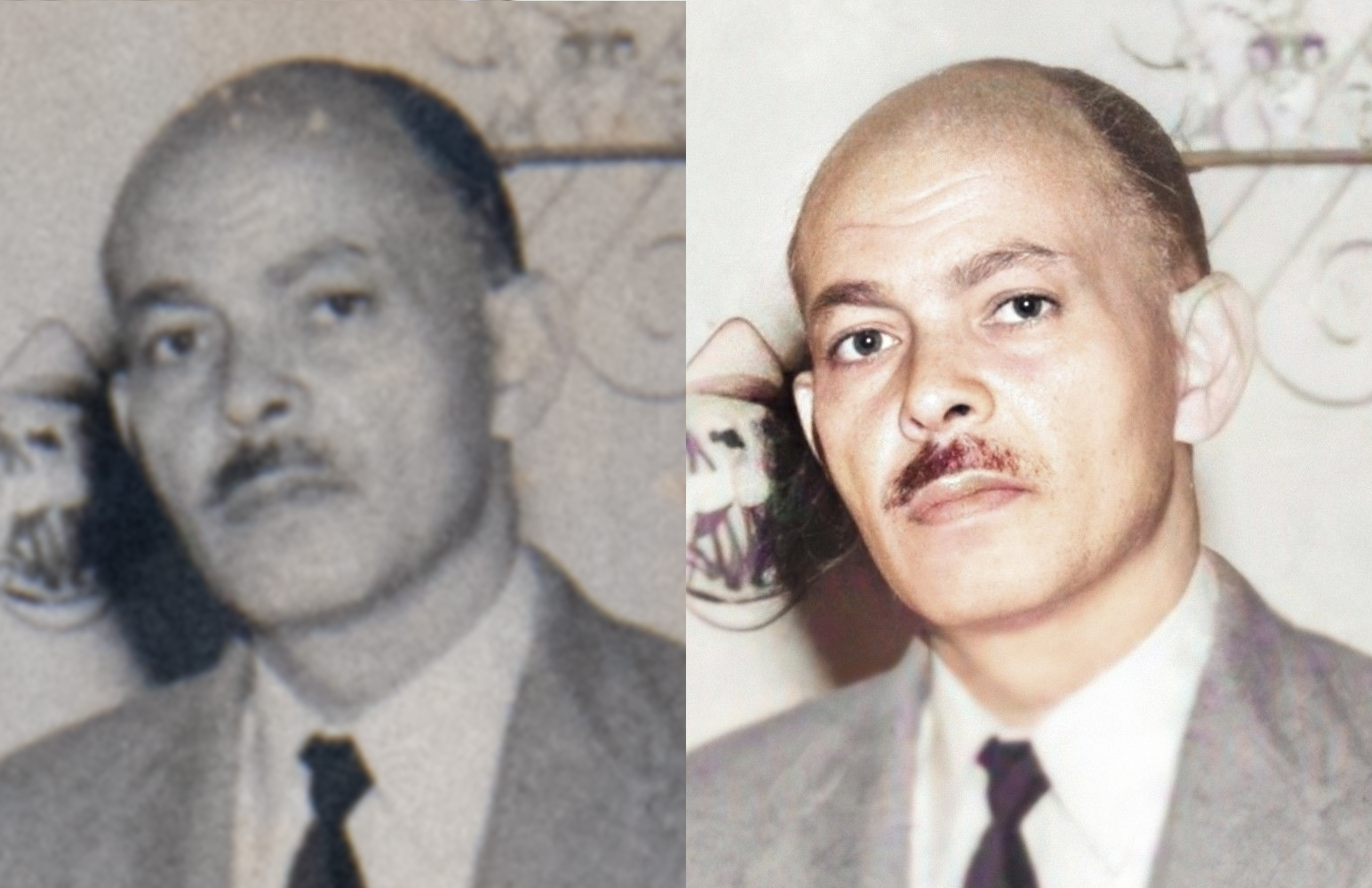}
   \caption[Source: https://en.wikipedia.org/wiki/Inpainting]{Image restoration using Artificial Intelligence.}
   \label{fig:restoration}
\end{figure}

\subsubsection*{When does applying AI make sense?}
Implementing an AI system makes sense when working with large amounts of data. In this large amount of data, the algorithm can approximate a generalization for a problem that can handle data which it has never seen before. Getting clean, balanced and unbiased data can sometimes be very difficult, though. The \textbf{Return on Investment (ROI)} is a value that is crucial when deciding over the implementation of an AI system. This means weighing up the cost and complexity of the implementation versus the benefit the AI can give while operating. Simpler tasks like for example regulating the heating of a house can simply be done autonomously without an AI operating in the background - the ROI of the AI solution might not be worthwhile here. Regarding more complex tasks, such as autonomous driving, the ROI gets dramatically better. Solving this task without the help of AI would seem infeasible.

Most tasks that can be completed with a data-defined pattern can be done with AI. Depending on the difficulty of the pattern it tries to approximate, it may be more efficient to manually program in all the rules for each case. As AI continues to advance, gathering sufficient data and implementing the algorithm is going to become simpler and simpler from a business perspective. The ROI for AI solutions will continue to ascent and AI will make its entrance into more and more sectors.

\subsection{Challenges regarding AI}
\subsubsection{Reliability} \label{reliability}
The further AI advances, the more complex and important tasks it will be able to take on. More important tasks also mean greater responsibility, especially as an increasing number of humans rely on that system. Therefore, \textbf{technical AI safety} becomes increasingly more important. While being able to empirically prove the accuracy or performance of an AI system in the lab, real-world scenarios always hold an unforeseen situation. This effect is called "data shift" and is an important problem to regard when trying to build a reliable AI system.\\

The technical safety of an AI can be divided into three areas: specification, robustness and assurance. Defining the right purpose of the system and trying hard to avoid ambiguities and side-effects caused by the way the system is designed is what falls into the area of specification. The robustness of a system is measured by its ability to withstand unwanted interference. This can either be done by proper training (see practical part of this work) or by error-correction and fallback mechanisms for recovery. The last area, assurance, is mainly about monitoring (see next section) and checking the system activity while in production.
\cite{reldeepmind}
\subsubsection{Interpretability}
To be able to use the power of AI for important decisions without blindly relying on it, we need to know based on what factors the AI decides.\\

Today, we mostly don't know how to interpret/explain the reasoning behind complex AI systems. That is why one of the problems we will probably have to work on is the interpretability of the algorithms. The term "\textbf{Explainable AI (XAI)}" refers to AI systems that possess some amount of "transparency", which means that humans can understand in at least some detail, why the algorithm came to the solution it did \cite{xai}. The challenge of creating XAI is to improve the "transparency" of the model to a level where it can prove useful to humans without compromising the performance of the model too much. Also, an "effective explanation" of the model can vary between different industries and users of the AI system. A developer, for example, would rather like to know possible areas of improvements, while a judge would rather want to know if the AI is making fair decisions \cite{aaasxai}.
\subsubsection{Moral Dilemma}
"Moral dilemmas are situations in which the decision-maker must consider two or more moral values or duties but can only honour one of them; thus, the individual will violate at least one important moral concern, regardless of the decision." \cite{md}. In moral dilemmas, because all choices are bad, deciding is difficult and relies heavily on personal biases. The question regarding AI is the following: Should we give AI systems the responsibility of making ethical decisions? How do we determine in which way an AI system reacts to moral dilemmas? To feel how difficult deciding becomes in dilemma situations, visit \url{https://www.moralmachine.net/} (see QR-code below).
\begin{figure}[H]
   \centering
   \includegraphics[width=.3\textwidth]{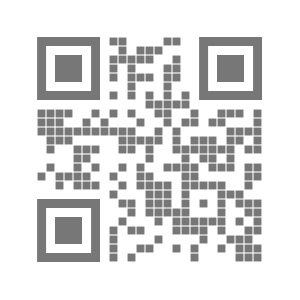}
\end{figure}
As you've probably seen, it can sometimes be very difficult, and as self-driving cars or other AI systems carry more responsibility, how we solve such situations becomes more important. \\

Another stage where moral questions have to be answered is when AI becomes sentient or even self-aware. When AI starts to be conscious, is it correct to just unplug it or delete it? Or even further, should the AI system have human or civil rights?\\

There are still a lot more questions regarding morality. For example, should AI be banned for war purposes like chemical and biological weapons since the Geneva Protocol? Is it morally reprehensible to further develop AI systems regarding all the future threats? We cannot and don't want to try to give any answers to these questions, but draw attention to the fact that they exist.

\subsubsection{\mbox{Energy and Computing Power Demand}}
In 2016, when AlphaGo defeated Lee Sedol in the game of Go, it used 1202 CPUs and 176 GPUs to compute its next move. To operate, it consumed about 1 megawatt, which, compared to Lee Sedol, is about 50'000 times more power \cite{alphago}. \\

Using such big amounts of energy is impractical, expensive and environmentally harmful. There is a general rule that the more complex the patterns to find or the task to solve, the more complex and deep your network should be. But this also comes at the cost of more computations and energy needed.\\

There have already been developed techniques to drastically reduce the size/"weight" of neural networks while keeping up the good performance. One of the techniques is called "Pruning", in which the researcher iteratively removes connections between neurons and retrains the network afterwards to achieve a minimum level of redundant connections in the network \cite{pruning}. Efficiency gains can also be achieved through tailored hardware, such as Google's TPUs (as explained in \ref{TPUs} under paragraph "Google"), as well as in the specialized hardware architectures developed all around the world, including at ETH Zurich and UZH.
\subsection{Outlook}
There are still many unresolved questions regarding the technology of AI. This is where we would like to leave the floor to the experts in the field, which is why we have conducted an interview with Prof. Dr. Martin Vechev, who leads the Secure, Reliable, and Intelligent Systems Lab (SRI) at ETH Zurich.\\

Q: \textit{Should AI be regulated more/less or is it just right and why? Are we on the way to losing control over AI? What are possible disadvantages of AI?}

A: \textit{I believe we are far away from losing control over AI. While a scenario where an Artificial
General Intelligence starts to improve itself in a runaway fashion is a popular topic for
Science Fiction novels and might become a concern in the far future, we are still very far
away from such a form of AI.
Therefore, I think the time to regulate AI more strictly has not yet come, as we simply do not
yet know what the AI systems of the future will look like and should avoid stifling innovation
by overly strict regulations.
When talking about individual tasks where AI is now used instead of other systems, it might
make sense to generally regulate how decisions are made. I don’t see a reason to apply
such regulations only to AI Systems and not the decision making system used before.
AI is like any tool, in that it has to be used correctly to produce good results. A danger might
be that it is harder to assess whether it is working well. The performance might vary
significantly depending on the circumstances. For example, if the dataset used to train and
evaluate a model is heavily biased in the same way, this will go undetected during testing but
might have unforeseen consequences once it gets applied in a real-world environment.
Therefore it is important to be careful and diligent when implementing AI solutions starting
with the gathering of data.}\\

Q: \textit{In what ways are AI systems exploitable and what can we do against it? (Adversarial attacks etc.)}

A: \textit{AI systems are exploitable in several ways. The training data might be biased, intentionally
or not, manipulated data can be injected into the training procedure and the resulting
behaviours later exploited, or the data evaluated by an AI system could be engineered to
mislead the system. These maliciously manipulated data points can fool AI systems that are
not trained to be robust to such attacks, while looking unchanged to human eyes. These
weaknesses each require different countermeasures.
Methods to analyse the vulnerability of an AI system to these adversarial examples and
training algorithms to defend against them are an active research topic that we are also
working on. Many methods exist to generate empirical or even deterministic robustness
guarantees, although those come at the cost of increased computational cost and reduced
accuracy.}\\

Q: \textit{Who is responsible for AI systems when they fail?}

A: \textit{Who will be responsible for a failing AI system is certainly an interesting question that will
most likely have to be decided by regulators.
In general, I believe it is unreasonable to expect that an AI system will be 100
under all (potentially unforeseen or maliciously engineered) circumstances, therefore
defining what a failure is, is critical to answer this question.
If you are referring to cases such as autonomous driving systems confusing a bright side
panel of a trailer or car with the sky and failing to detect an obstacle, the entity providing the
system is probably responsible for the failure of the AI system. However, this is not the same
as responsibility for a potentially resulting accident. An observant driver who could have
intervened, society in general exaggerating the ability of such a system or a regulator not
defining appropriate certification requirements might also all be partially responsible.
}\\

Q: \textit{What’s the outlook on the optimization of the power consumption of AI algorithms? Is there a lot of room for optimization? AlphaGo consumed about 1 megawatt, which, compared to Lee Sedol, is about 50'000 times more power.}

A: \textit{Energy efficiency is an important question in AI research and there are significant efforts to
reduce the energy and hardware requirements to enable the use of these systems for
example on mobile phones. While there probably is still significant room for optimization, a
factor of 50,000 seems unlikely.
The AlphaGo example is also an extreme case demonstrating what is possible if efficiency
constraints are ignored. Considering that the successor to AlphaGo, AlphaGo Zero, is
estimated to have an Elo of over 5000, corresponding approximately to losing only 1 out of
1000 games against the best human go players, using more power might also be
acceptable.
}\\

Q: \textit{Is the brain efficient? (in the sense that it only can be simulated by a thing bigger and more complicated than itself)}

A: \textit{It is almost universally true that it requires a more complex system to fully simulate the
behaviour of a system, this is called computational irreducibility, while there are some
exceptions to this behaviour, none have yet been found for the brain. The question is,
whether you want to actually simulate a human brain, or solve the same tasks. The latter
seems like the more interesting problem and makes the question harder to answer. If you
think about complex mathematical operations the human brain is orders of magnitudes
slower and less efficient than even a simple calculator, while AI systems can not replicate
other feats. Therefore, I will go with: The brain is probably efficient for some tasks but not for others.}\\

Q: \textit{Try to estimate the impact of AI in the workplace in the short term (5-10 years) and in the long term (50+ years). Will humans become obsolete in the workplace?}

A: \textit{I predict the biggest short term impact of AI on the workplace to be in the transportation field.
Autonomous driving in combination with electrification has the potential to revolutionize both
cargo and personal transportation.
To make predictions about the far future would only be guessing. Everything from regulators
clamping down on AI so heavily as to make it unattractive up to true Artificial General
Intelligence is possible. While I don’t believe that humans will become obsolete in the
workplace, their work might shift to tasks focused more around human interaction.
}\\

Q: \textit{Is the creation of Artificial Intelligence the most important event in human history so far?}

A: \textit{Achieving Artificial General Intelligence could become the most important event in human
history, but until then I don’t believe so.}\\

Q: \textit{How far are we from Artificial General Intelligence and a Superintelligence?}

A: \textit{It does not look like the development of an AGI or Superintelligence is imminent, but whether
it will take 20 or 200 years, if it ever happens, is difficult to say.}\\

\section{Appendix}
\subsection{Reasons for the choice of topic}
AI could unravel to be the most important invention of humankind of our time or even of all time. Because of this and the rapid development during this decade it was an easy decision for us to choose AI as the subject of our Matura Thesis. We believe that the future lies in AI and therefore it is worthwhile to have engaged with it. As potential computer science students we could imagine a specialization towards AI in the later course of our studies and thus a job in this field. So far only the tip of the iceberg has been scratched, the field of AI will take a big step in the direction of progress during our lifetimes and we strive to contribute to this development.
\subsection{Acknowledgements}
We would first like to thank our thesis advisor Prof. Dr. Delbrück Tobias of the Institute of Neuroinformatics. With any problems or questions we encountered he guided and encouraged us to do the better even when it was not easy. He also gave us access to the needed equipment for this project, which we profoundly appreciate.

Of course, we also thank the PhD students Yuhuang Hu and Shasha Guo, whom Tobias Delbrück provided for additional help. They were very responsive and we therefore could always rely on quick support for smaller questions.

We would also like to acknowledge the work of Michael Bucher and Christopher Latkoczy from Kantonsschule Stadelhofen as our school intern advisors and proof-readers of this thesis, we are gratefully indebted to their valuable suggestions on this thesis.

Finally, we must express our deepest gratitude to our family and close friends for providing us with unfailing support and continuous encouragement throughout the process of researching and writing this thesis. This accomplishment would not have been possible without them. Our sincerest gratitude to you.
\end{multicols*}
\pagebreak
\bibliographystyle{unsrt}
\begingroup
\raggedright
\bibliography{uni}
\pagebreak
\listoffigures
\endgroup
\end{document}